%% file: main_arxiv.tex
\documentclass[runningheads]{llncs}

\usepackage{eccv}

\usepackage{eccvabbrv}
\usepackage{float}
\usepackage{graphicx}
\usepackage{booktabs}

\usepackage{adjustbox}
\usepackage{multirow}
\usepackage{amssymb}
\usepackage{amsmath}
\usepackage{makecell}
\usepackage{subcaption}
\usepackage{pgfplots}
\pgfplotsset{compat=1.18}

\usepackage{tikz}
\usetikzlibrary{patterns}
\usetikzlibrary{pgfplots.groupplots} 

\usepackage{pifont}
\newcommand{\cmark}{\ding{51}}%
\newcommand{\xmark}{\ding{55}}%

\usepackage[accsupp]{axessibility}  

\usepackage{hyperref}

\usepackage{orcidlink}

\makeatletter
\renewcommand{\@fnsymbol}[1]{\ensuremath{\ifcase#1\or \star\or \dagger\or \ddagger\or \mathsection\or \mathparagraph\else\@ctrerr\fi}}
\makeatother

\begin{document}

\title{Defect-Aware Hybrid Prompt Optimization for Zero-Shot Multi-Type Anomaly Detection and Segmentation} 

\titlerunning{Defect-Aware Hybrid Prompt Optimization}

\author{Nadeem Nazer\inst{1,2}\thanks{These authors contributed equally.} \and
 Hongkuan Zhou\inst{1,3}\textsuperscript{$\star$}\textsuperscript{$\dagger$} \and
 Lavdim Halilaj\inst{1} \and
 Ylli Sadikaj\inst{4} \and Steffen Staab\inst{3,5}
}
\authorrunning{N. Nazer, H. Zhou, et al.}

\institute{Corporate Research, Robert Bosch GmbH, Renningen, Germany \\
\textsuperscript{$\dagger$} \email{Hongkuan.Zhou@de.bosch.com}
\and
Otto-von-Guericke-University Magdeburg, Magdeburg, Germany \and
Institute for Artificial Intelligence, University of Stuttgart, Stuttgart, Germany \and
Faculty of Computer Science, UniVie Doctoral School Computer Science, University of Vienna, Vienna, Austria \and
University of Southampton, Southampton, UK \\
}
\definecolor{network-blue}{RGB}{165, 192, 221}
\definecolor{light-yellow}{RGB}{238, 233, 218}
\definecolor{light-green}{RGB}{129, 184, 113}
\definecolor{light-red}{RGB}{242, 182, 160}
\definecolor{light-blue}{RGB}{124, 150, 171}
\definecolor{light-orange}{RGB}{255,147,0}
\definecolor{light-purple}{RGB}{150,115,166}
\definecolor{dark-green}{RGB}{85, 124, 86}
\definecolor{dark-red}{RGB}{217, 22, 86}
\definecolor{purple}{RGB}{155, 126, 189}
\definecolor{dark-purple}{RGB}{59, 30, 84}
\definecolor{dark-blue}{RGB}{0,91,159} 

\maketitle

\input{sec/0_abstract}
\input{sec/1_intro}
\input{sec/2_related_works}
\input{sec/3_preliminary}
\input{sec/4_methodology}
\input{sec/5_experiments}
\input{sec/6_results}
\input{sec/7_conclusion}
%
\bibliographystyle{splncs04}
\bibliography{main}
\appendix
\input{sec/X_suppl}
\end{document}

%% file: sec/0_abstract.tex
\begin{abstract}
Recent vision-language models (VLMs) like CLIP have shown impressive anomaly detection performance under significant distribution shift by utilizing high-level semantic information through text prompts.
However, these models often overlook fine-grained defect cues, e.g., hole, cut, or scratch, that are essential for understanding the anomaly's nature. Moreover, the modality gap between images and text can lead to subtle visual evidence being poorly captured in textual descriptions.
To address the gap, we enhance the representation of ``abnormal'' with structured semantics, bridging coarse anomaly signals and fine-grained defect categories. We propose a hybrid prompting mechanism that combines human-readable descriptions of defect types with learnable token embeddings. Building on these ideas, we introduce DAPO, a \textbf{D}efect-\textbf{a}ware \textbf{P}rompt \textbf{O}ptimization framework for zero-shot multi-type and binary anomaly detection and segmentation under distribution shift. DAPO aligns anomaly-relevant visual features with their corresponding textual semantics by learning hybrid defect-aware prompts that combine fixed textual anchors with trainable token embeddings.
We conducted experiments on public benchmarks (MPDD, VisA, MVTec-AD, MAD, and Real-IAD) and an internal dataset.
The results suggest that compared to the baseline models, DAPO achieves a 3.6\% average improvement in AUROC and average precision metrics at the image level under distribution shift, and a 5.2\% average improvement in AUROC and F1 when localizing novel anomaly types under zero-shot settings.
\end{abstract}

%% file: sec/1_intro.tex
\section{Introduction}
\label{sec:intro}
\begin{figure}
    \centering
    \includegraphics[width=\linewidth]{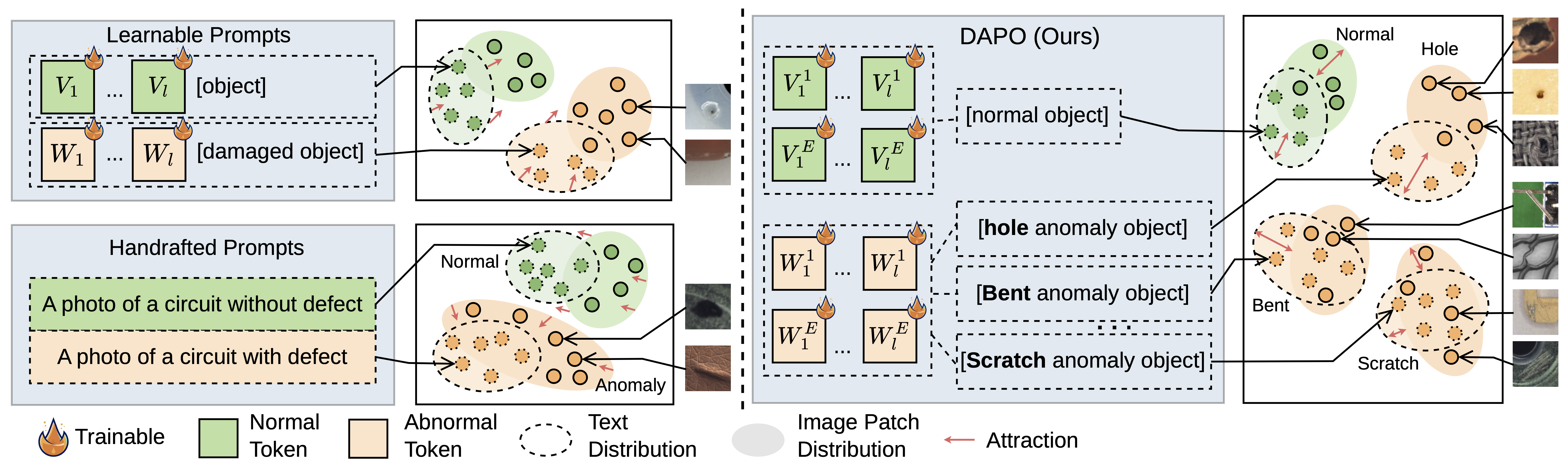}
    \caption{Motivation. Existing CLIP-based prompting often yields coarse normal–abnormal separation, where different defect types collapse into overlapping anomaly representations. DAPO uses compositional defect-aware prompts to align visual cues with defect semantics, producing better-separated representations across defect types while avoiding manual prompt enumeration. In particular, the shared learnable context tokens capture a type-agnostic abnormal context that transfers across defect categories, while the defect-name anchors provide type-specific semantics.}
    \label{fig:motivation}
    \vspace{-1.5em}
\end{figure}
Visual inspection is a crucial task in manufacturing, ensuring that the products are free of defects before reaching customers. Manual efforts in such inspections incur high labor costs and attention. With advancements in deep learning and computer vision techniques, these complex processes are increasingly being automated, reducing manual involvement and enabling high-precision identification of defects. Therefore, visual inspection through anomaly detection (AD) and segmentation is essential in identifying and localizing defective regions.


Most conventional approaches adopted one-class or unsupervised anomaly detection \cite{zolfaghari2022unsupervised, yang2020dfr, roth2022towards, defard2021padim, deng2022anomaly}, where they are trained solely on the volumes of normal (non-defective) images and classify any deviation from the learned distribution as an anomaly. Although such models perform well on their training distribution, the domain shift in industrial settings is large, so that these models often need to be trained separately on different domains and frequently retrained. Consequently, recent work increasingly leverages large-scale pre-trained VLMs (e.g., CLIP~\cite{radford2021learning}), using their broad priors as an adaptable foundation that mitigates data scarcity and reduces extensive domain-specific annotation.

Despite their potential, existing approaches remain limited to coarse-grained anomaly detection, i.e., identifying whether an image is abnormal without differentiating defect types. This limitation matters in practice for two reasons. 1) type-level semantics provide an inductive bias that aligns local features with defect-specific prototypes, improving robustness under distribution shift~\cite{sadikaj2025multiads}; 2) focusing on the nature of anomalies such as stains, holes, or missing components is essential for industrial operators to trace root causes and implement preventive measures. 
Prior work, such as MultiADS~\cite{sadikaj2025multiads}, attempts to incorporate defect-type knowledge via a curated knowledge base and hand-crafted prompts, but this process is time-consuming and difficult to scale as new defect categories emerge. Prompt learning offers an appealing alternative and has been successfully used in works like AnomalyCLIP~\cite{zhou2023anomalyclip}. 
However, relying solely on learnable (soft) prompt tokens can drift away from the natural-language manifold on which CLIP generalizes well thanks to its large-scale pretraining, thereby weakening zero-shot transfer ability when tested on the target domain.

To overcome these challenges, we propose DAPO, \textbf{D}efect-\textbf{a}ware \textbf{P}rompt \textbf{O}ptimization, where we learn hybrid prompts for each defect type, which consist of a set of shared learnable defect token embeddings together with the literal tokens of specific defect-type descriptions in textual format (e.g., $[W_1], \ldots, [W_l]$ [``bent''] [``defect'']). Sharing learnable tokens across defect categories enables prompt construction for novel defect types by reusing the same learned abnormal tokens while simply swapping in the textual description of the new defect. The main contributions of this work are as follows:

\begin{itemize}
    \item We propose DAPO, a defect-aware prompt optimization framework that composes hybrid prompts from fixed textual anchors and a small set of shared learnable defect tokens reused across defect types.
    \item We design a prompt composition and optimization scheme that explicitly bridges the coarse anomaly concept and fine-grained defect semantics, improving image-text alignment. 
    \item We conduct comprehensive experiments on five public industrial benchmarks and an internal semiconductor dataset under significant distribution shifts, demonstrating consistent gains on both zero-shot multi-type anomaly segmentation and binary anomaly detection. Our code is publicly available.\footnote{\url{https://github.com/boschresearch/defect-aware-prompt-optimization}}
\end{itemize}

%% file: sec/2_related_works.tex
\section{Related Works}
\subsection{Unsupervised Unimodal Approaches} 

Such approaches model normal distributions from training data, treating deviations as anomalies. Reconstruction-based methods~\cite{venkataramanan2020attention, liu2021unsupervised, yang2020dfr}, leverage architectures of GAN~\cite{goodfellow2020generative}, Variational AEs~\cite{kingma2013auto}, and AEs~\cite{lecun1989generalization} to find anomalies. Other approaches, such as PatchCore~\cite{roth2022towards}, PaDIM~\cite{defard2021padim}, and SPADE~\cite{cohen2005sub}, adopt representational or memory-bank-based strategies by storing normal image features and measuring distances between test samples and those reference features using different distance measurement techniques. STPM~\cite{zolfaghari2022unsupervised} and RD4AD~\cite{deng2022anomaly} utilize a knowledge distillation-based student-teacher framework for AD, using feature matching performed utilizing a pre-trained classification model. Anomalies are detected by measuring feature discrepancies between the student and teacher networks. However, these methods face several limitations, including dependence on training data and a one-class classification focus. These challenges motivated the development of zero-shot approaches.

\subsection{Zero-Shot Anomaly Detection} 
Zero-Shot Anomaly Detection is gaining attention due to advancements in large-scale vision-language models (VLMs), which have wide application in robotics~\cite{zhou2023language, 10685120, 10934975} and autonomous driving~\cite{10531702, fm4su}.
CLIP, in particular, is serving as a widely adopted backbone for this task because of its strong zero-shot visual recognition capability. 
Several approaches emerged, adapting CLIP for AD, such as WinCLIP~\cite{jeong2023winclip}, which uses manually crafted normal and abnormal state prompts with a window-based segmentation approach.
AprilGAN~\cite{chen2023april} introduces additional trainable linear layers to adapt CLIP for AD. 
Drawing inspiration from prompt learning success in NLP, methods such as CoOp~\cite{zhou2022learning} and CoCoOp \cite{zhou2022conditional} learn task-specific prompts. 
AnomalyCLIP~\cite{zhou2023anomalyclip}, replaces manually defined prompts with learnable object-agnostic tokens to form a general text representation for normal and abnormal classes. 
Similarly, SimCLIP~\cite{deng2024simclip} uses implicit prompt tuning to refine text embeddings for better alignment with CLIP's patch features, and AdaCLIP~\cite{cao2024adaclip} combines static and dynamic prompts in a hybrid learning framework. 
More recently, CLIPSAM~\cite{li2025clipsam} introduced a novel multi-modal approach by integrating CLIP with the Segment Anything Model (SAM)~\cite{kirillov2023segment} for enhanced anomaly segmentation and detection capabilities. 
However, all these methods focus on learning or crafting prompts that can capture only coarse-grained features, failing to identify anomalies with different specific types. 
FiLo~\cite{gu2024filo} introduces fine-grained anomaly descriptors generated by LLMs and adaptively learned textual templates, replacing general ``abnormal'' semantics with category-specific descriptions, but it does not learn a shared learnable component reused across defect types.
FAPrompt~\cite{zhu2024fine} learns compound prompts for fine-grained feature extraction, yet it has no explicit defect-name anchor and is formulated for binary AD rather than explicit defect-type segmentation.
Recent work such as LTOAD~\cite{LTOAD} explores the use of multiple learnable concept-specific prompts, targeting Online AD by updating prompts at test time to handle long-tailed distributions, whereas Anomaly-OV~\cite{xu2025anomalyov} is a multimodal large language model (MLLM)-based detection-and-reasoning framework. Both are complementary to, but not directly comparable with, our lightweight frozen-CLIP zero-shot setting. Furthermore, a major limitation across the aforementioned prompt-learning designs is their formulation for binary normal-vs-anomalous scoring. While MultiADS~\cite{sadikaj2025multiads} represents a representative attempt at multi-defect type detection, it relies heavily on extensive hand-crafted prompts, which are difficult to scale.

Unlike AdaCLIP's global-vs-dynamic (image-conditioned) prompts, DAPO composes a fixed defect-name anchor with a set of \emph{shared} learnable tokens. Unlike FiLo and FAPrompt, it supports zero-shot instantiation of previously unseen defect types by simply swapping the textual anchor, without retraining or updating an external descriptor set.
Our approach mitigates these limitations through hybrid defect-aware prompting, where learnable tokens are composed with fixed textual anchors, and through local alignment objectives that couple defect semantics to fine-grained visual evidence, reducing the discrepancy between visual and textual representations.

%% file: sec/3_preliminary.tex
\section{Preliminaries}
We provide the formal definition of three anomaly detection tasks addressed in this work: (1) binary anomaly detection, (2) binary anomaly segmentation under distribution shift, and (3) zero-shot multi-type anomaly segmentation. Note that the task of binary anomaly detection/segmentation under distribution shift is commonly referred to as \emph{zero-shot} anomaly detection/segmentation in the literature. However, we argue that this term is \emph{misleading}, as the objective is not to locate new types of anomalies, but to distinguish between normal and anomalous samples when applied to previously unseen products, which should be seen as a distribution shift.

\paragraph{Common setup.}
Let $\mathcal{D}_{\text{train}} = \{(\mathbf{x}_i, \cdot)\}_{i=1}^{N_1}$ and $\mathcal{D}_{\text{target}} = \{(\mathbf{x}_i, \cdot)\}_{i=N_1+1}^{N_1+N_2}$ be source and target datasets whose images are drawn from different input distributions: $\mathbf{x}_i \sim \mathcal{P}(\mathcal{X})$ for $i\in\{1,\dots,N_1\}$ and $\mathbf{x}_i \sim \mathcal{P}(\mathcal{X'})$ for $i\in\{N_1+1,\dots,N_1+N_2\}$, with $\mathcal{P}(\mathcal{X}) \nsim \mathcal{P}(\mathcal{X'})$ and each image $\mathbf{x} \in \mathbb{R}^{H \times W \times 3}$. A model $\hat{f}(\cdot)$ is trained solely on $\mathcal{D}_{\mathrm{train}}$ and, at test time, adapted to $\hat{f'}(\cdot)$ using auxiliary target-domain metadata, such as the set of relevant defect types. The three tasks differ in their label space and prediction target.

\paragraph{Binary anomaly detection under distribution shift.}
Each image is labeled with $y \in \{0,1\}$ indicating anomaly presence. For each $\mathbf{x}\in \mathcal{D}_{\mathrm{target}}$, the detector $\hat{f'}(\cdot)$ classifies it as normal or anomalous.

\paragraph{Binary anomaly segmentation under distribution shift.}
Each image is paired with a binary mask $\mathbf{y} \in \{0,1\}^{H \times W}$ indicating the presence of anomalies at each pixel. For each pixel of $\mathbf{x}\in \mathcal{D}_{\mathrm{target}}$, the model $\hat{f}^\prime(\cdot)$ classifies it as normal or anomalous.

\paragraph{Zero-shot multi-type anomaly segmentation.}
In \(\mathcal{D}_{\mathrm{train}}\), each image is paired with a mask \(\mathbf{Y}\in\{0,1\}^{H\times W\times(K_1+1)}\) encoding one normal class and \(K_1\) defect types \(\{d_1,\dots,d_{K_1}\}\); in \(\mathcal{D}_{\mathrm{target}}\), each image is labeled with a mask over one normal class and a possibly different set of \(K_2\) defect types. Given the target defect types as external knowledge, $\hat{f'}(\cdot)$ predicts, for every \(\mathbf{x}\in\mathcal{D}_{\mathrm{target}}\), a mask \(\hat{\mathbf{Y}}\in\{0,1\}^{H\times W\times(K_2+1)}\) that locates anomalies of each defect type.

%% file: sec/4_methodology.tex
\section{Methodology}
\label{ch:methodology}

\begin{figure*}[t]
    \centering
    \includegraphics[width=\textwidth]{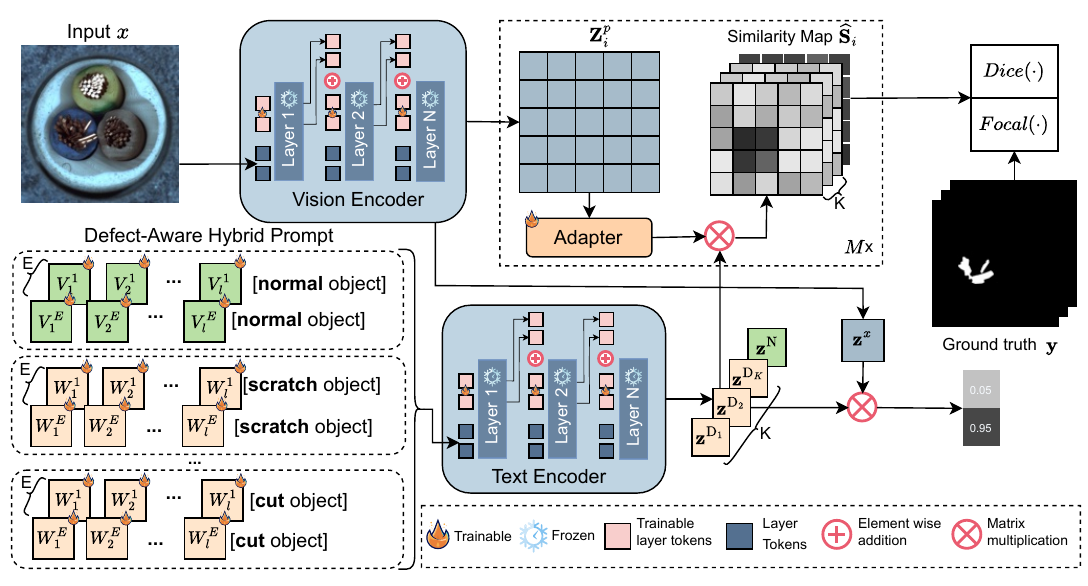}
    \caption{The training pipeline of DAPO. We learn compositional defect-aware prompts by combining shared learnable context with defect-name anchors, and align them with global and patch-level visual features via global and local objectives. Only prompt tokens and lightweight adapters are updated; CLIP encoders remain frozen.}
    \label{fig:architecture}
\end{figure*}

We propose DAPO, a CLIP-based framework for anomaly detection and defect-type localization. DAPO learns compositional defect-aware prompts and aligns them with both global image features and patch-level features through global and local objectives (Fig.~\ref{fig:architecture}). At test time, prompts for unseen defect descriptors are instantiated by replacing the defect-description anchor $\langle D\rangle$.

\subsection{Defect-Aware Prompt Design}
\label{sec:prompt-design}
We propose a defect-aware learnable prompts, which contain both human-read\-able words and learnable embeddings that can effectively capture abnormal and normal states (cf. Figure~\ref{fig:architecture}). Specifically, a prompt for the abnormal state is composed as:
\begin{equation}
    \mathbf{P}^{\text{D}} = [W_1][W_2]...[W_l]\langle D\rangle[\text{anomaly}][\text{object}] \text{,}
    \label{eq:defect-prompt}
\end{equation}
where $[W_i], i\in\{1,2,...,l\}$ are learnable context tokens that model a shared, type-agnostic context for abnormality. $\langle D\rangle$ is a placeholder for the defect descriptor, such as \{``bent'', ``contamination'', \dots\}. The tokens [anomaly] and [object] are fixed literals corresponding to the words ``anomaly'' and ``object''.

A key design choice is to share the learnable tokens ${[W_i]}$ across all defect types, yielding a single set of parameters that learns a defect-aware context composable with different descriptors: the explicit defect word $\langle D\rangle$ injects type-specific semantics (e.g., ``scratch'', ``crack''), while the shared tokens provide transferable anomaly knowledge that improves generalization across defect categories. The fixed anchor $[\text{anomaly}]$ grounds the prompt in CLIP’s pre-trained semantic space, and $[\text{object}]$ preserves object-level context while allowing the learnable tokens to focus on defect cues that generalize across object categories~\cite{zhou2023anomalyclip}. 
Similarly, the normal prompt can be represented as:
\begin{equation}
    \mathbf{P}^{\text{N}} = [V_1][V_2]...[V_l][\text{normal}][\text{object}] \text{.}
    \label{eq:normal-prompt}
\end{equation}
where $[V_i], i\in\{1,2,...,l\}$ are learnable context tokens for normal state. Building on this foundation, to further enhance the representation capability, for one normal and $K$ abnormal states of different defect types, we generate $E$ prompts, each capturing different aspects of the anomaly space. For any given defect type, all $E$ prompts are utilized, meaning that a single defect type (e.g., ``scratch'') is described through $E$ different learnable contextual embeddings. The prompt set can be written as: 
\begin{align}
    \mathcal{B} & =  \{\mathbf{P}^{\text{N}}_1, \mathbf{P}^{\text{N}}_2, \dots \mathbf{P}^{\text{N}}_{E}\} \text{,}
      \nonumber
    \\
    \mathcal{A}_k& =  \{\mathbf{P}^{\text{D}_k}_1, \mathbf{P}^{\text{D}_k}_2, \dots \mathbf{P}^{\text{D}_k}_{E}\} 
    \text{,}
    \label{eq:prompt set}
\end{align}
where $\mathcal{B}$ is the prompt set for the normal state, while $\mathcal{A}_k$ is the prompt set for the abnormal state of the $k$-th defect.

\subsection{Prompt Optimization}
\label{ch:learning_prompt}
Instead of fine-tuning the parameters of the network, we optimize the prompt set we defined in Equation~\eqref{eq:prompt set}. In the following section, we discuss the encoding process of images and texts and the loss function design for local and global optimization.

\subsubsection{Image Encoding}
Given an input image $\mathbf{x} \in \mathbb{R}^{H \times W \times 3}$, the image encoder generates intermediate image patch embeddings at $M$ encoding stages or layers and a single global image embedding $\mathbf{z}^{x}$ in the shared vision-text semantic space $\mathcal{Z}$. For each stage $i \in \{1,2,\dots,M\}$, a linear adapter transforms the corresponding patch embeddings into the vision-text semantic space $\mathcal{Z}$. We denote these adapted embeddings as $\mathbf{Z}^{p}_{i} \in \mathbb{R}^{C \times H^\prime \times W^\prime}$. $H^\prime \times W^\prime$ are the resolution of image patch embeddings. Formally, the output of the image encoder is defined by:

\begin{equation}
(\mathbf{z}^{x}, \mathbf{Z}^{p}_1, \dots, \mathbf{Z}^{p}_{M}) = f_{\theta}(\mathbf{x})\text{,}
\end{equation}
where $f_{\theta}(\cdot)$ is the parameterized image encoder function.

\subsubsection{Text Encoding}
For each prompt $\mathbf{P}\in \mathcal{A}_k$, the text encoder $f_\lambda(\cdot)$ encodes the prompt to a text embedding. The embedding representing the $k$-th defect state is
\begin{equation}
    \mathbf{z}^{\text{D}_k} = \frac{1}{E} \sum_{\mathbf{P}\in\mathcal{A}_k}f_\lambda(\mathbf{P}), k \in\{1,...,K\}
\end{equation}
Similarly, the embedding for the normal state is 
\begin{equation}
    \mathbf{z}^{\text{N}} = \frac{1}{E} \sum_{\mathbf{P}\in\mathcal{B}}f_\lambda(\mathbf{P}) \text{.}
\end{equation}
In total, we have $K+1$ embeddings $\{\mathbf{z}^\text{N}, \mathbf{z}^{\text{D}_1}, ...,\mathbf{z}^{\text{D}_K}\}$ represent one normal state and $K$ abnormal state. 

Following prior work on progressive prompt tuning, we adopt a progressive prompt injection strategy to stabilize optimization when inserting learnable tokens into both visual and textual encoders.

\subsection{Loss Function Design}
Our loss function consists of two components: The global loss uses cosine similarity to align the global image embedding with its corresponding normal or abnormal state, capturing high-level semantic differences between samples. This is achieved by calculating the cosine similarity between the global image embedding $\mathbf{z}^x$ with $\mathbf{z}^\text{N}$ and the averaged defect-aware prompts, $\frac{1}{K}\sum_{k=1}^{K}\mathbf{z}^{\text{D}_k}$ to get the similarity score $\mathbf{s} \in \mathbb{R}^2$. The global loss is defined by 
\begin{equation}
    \mathcal{L}_\text{global} = CE(softmax(\mathbf{s}), y)\text{,}
\end{equation}
where $CE(\cdot)$ represents the cross-entropy calculation.

The local loss aligns image patch embeddings with one of the $K+1$ states at a fine-grained spatial resolution.  Each adapted image patch embedding $\mathbf{Z}_i^p$ is compared with $K+1$ text embeddings to get the similarity maps. Since we choose image patch embeddings at $M$ different layers, we get $M$ different similarity maps $\hat{\mathbf{S}}_i\in \mathbb{R}^{(K+1) \times H^\prime \times W^\prime}$. This allows the model to localize and differentiate between defect types and normal regions at a finer spatial resolution. 
The local loss is:

\begin{align}
\label{eq:local_loss}
    \mathcal{L}_\text{local} = \frac{1}{M}\sum_{i=1}^{M} Focal(UP(\hat{\mathbf{S}}_{i}), \mathbf{Y}) +  Dice(UP(\hat{\mathbf{S}}_i[0]), \mathbf{Y}[0]) \notag \\ + Dice(\mathbf{1} - UP(\hat{\mathbf{S}}_i[0]), \mathbf{1}-\mathbf{Y}[0]) \text{,}
\end{align}
where $Focal(\cdot)$, and $Dice(\cdot)$ are focal~\cite{lin2017focal} and dice~\cite{li2019dice} losses, respectively. Focal loss is designed to address the class imbalance problem, which is especially relevant in AD, where anomalous samples are significantly fewer than normal ones. Dice loss encourages the model to learn accurate decision boundaries by measuring the overlap between the predicted and ground truth segmentation mask. $UP(\cdot)$ is the upsampling function to enlarge the resolution to the original image size $H \times W$.  
$\hat{\mathbf{S}}_i[0]$ represents the predicted probability map for the normal state at stage $i$. $\mathbf{Y}\in\mathbb{R}^{(K+1)\times H \times W}$ represents the ground truth of the multi-defect type segmentation and $\mathbf{Y}[0]$ is the layer for the normal part. 
The final loss is a weighted combination of the global and local losses:
\begin{equation}
\label{eq:loss}
    \mathcal{L}_\text{total} = \mathcal{L}_\text{global} + \lambda\mathcal{L}_\text{local}
\end{equation}
where $\lambda$ is a hyperparameter used for balancing.
It is important to note that the underlying feature spaces of both text and vision modalities are pre-trained to capture image-text semantics, rather than anomaly semantics. This motivates refining the embedding space to better align the modalities within the anomaly detection domain, as described in the following section.

\subsection{Training and Inference}
At inference, we instantiate $(K'{+}1)$ prompts (normal plus the target defect descriptors), aggregate the similarity maps across the $M$ stages, and obtain (i) a binary anomaly map as $UP(1-\tilde{\mathbf{S}}[0])$ and (ii) a multi-type mask by pixel-wise argmax over the $(K'{+}1)$ channels.

%% file: sec/5_experiments.tex
\section{Experiments}
In this section, we describe the experimental settings and benchmarks, define the evaluation metrics, and analyze the performance of DAPO through comparisons, ablations, and qualitative examples.

\subsection{Datasets}
Five public industrial datasets are used: MVTec-AD~\cite{bergmann2019mvtec}, VisA~\cite{zou2022spot}, MPDD~\cite{jezek2021deep}, MAD~\cite{zhou2023pad}, and Real-IAD~\cite{wang2024real}, as well as an internal semiconductor (ASIC) inspection dataset. 
Our internal dataset consists of Application-Specific Integrated Circuit (ASIC) inspection images collected from a real manufacturing line. Unlike public benchmarks that span multiple object categories and defect taxonomies, this dataset contains one product type and a single defect type (contamination), with pixel-level annotations of contaminated regions. Samples of this dataset are exemplified in Figure~\ref{fig:sub_b}. We evaluate both image-level anomaly detection and pixel-level segmentation and use this dataset as a focused case study to assess performance under realistic acquisition conditions.
For datasets with defect-type annotations, we report multi-type anomaly segmentation results using the dataset-provided defect labels.

\subsection{Implementation Details}
We use a frozen OpenCLIP~\cite{ilharco2021openclip} ViT-L/14@336 backbone with OpenAI pre-trained weights, and optimize only the learnable prompts and $M$ linear adapters while keeping both the vision and text encoders frozen throughout training. These trainable components constitute only $\sim$0.15\% of the model, allowing training in $\sim$25 min on MVTec-AD and inference in $\sim$55 min on the large-scale Real-IAD dataset on a single NVIDIA H200 GPU. Following a source$\rightarrow$target transfer protocol, we train on MVTec-AD and evaluate on the remaining benchmarks, and additionally train on VisA to evaluate on MVTec-AD. Input images are resized to $518{\times}518$ and normalized with the standard CLIP preprocessing pipeline. We optimize with Adam~\cite{kingma2014adam} (learning rate $10^{-3}$, batch size $8$) for $5$ epochs, implemented in PyTorch~2.7.0. We extract patch features from layers $\{6, 12, 18, 24\}$ ($M{=}4$) and set the prompt length $l{=}5$, the number of prompt instances $E{=}10$, and the local-loss weight $\lambda{=}4.0$. We initialize the learnable context tokens within CLIP's embedding space, matching its token-embedding mean $\mu$ and standard deviation $\sigma$, which converges faster and yields better results than random initialization; empirically, performance peaks within the first one or two epochs, with later epochs tending to overfit the source distribution. 

\subsection{Evaluation Metrics}
To evaluate the performance, we use the area under the receiver operating characteristic curve (AUROC), average precision (AP) for AD. For segmentation, we use pixel-level AUROC and the area under per-region overlap (AUPRO). For multi-type segmentation, we use AUROC, AP, and F1 by macro-averaging.
In addition, for deployment-oriented analysis on the internal dataset, we report operating-point metrics derived from the confusion matrix, including true positive rate (TPR) and false positive rate (FPR), where $\mathrm{TPR}=\frac{\mathrm{TP}}{\mathrm{TP}+\mathrm{FN}}$ and $\mathrm{FPR}=\frac{\mathrm{FP}}{\mathrm{FP}+\mathrm{TN}}$.
On the public benchmarks, all baseline numbers are \emph{reported} from their original papers or, where unavailable under our protocol, from MultiADS~\cite{sadikaj2025multiads}, and DAPO is evaluated under the same source$\rightarrow$target protocol; on the internal ASIC dataset, AnomalyCLIP and MultiADS are \emph{reproduced} by us using the authors' released code.

%% file: sec/6_results.tex
\section{Results}
We evaluate our approach on two primary tasks: (1) Binary Anomaly Detection and Segmentation under distribution shift. (2) Multi-Type Anomaly Detection, a task to evaluate fine-grained localization proposed in MultiADS~\cite{sadikaj2025multiads}.

\subsection{Multi-type Anomaly Segmentation}
We first evaluate our primary contribution: using prompt-learning for zero-shot multi-type anomaly segmentation. We compare our method with the handcrafted prompt baseline, MultiADS~\cite{sadikaj2025multiads}. The performance comparison is detailed in Table~\ref{tab:Multi-Type}. 
To the best of our knowledge, we are among the first to apply a prompt-learning approach to the zero-shot multi-type anomaly segmentation task. Our approach consistently outperforms MultiADS, measured by the AP metric, and achieves comparable or superior pixel-level F1 scores. The performance margin is the largest in MPDD with a +7.2\% on F1-score and a +6.0\% on MAD-Real datasets, showing the strength of our prompt-learning approach for localizing visually present anomalies such as ``scratch'' or ``crack''. However, our method shows a slight decrease in the VisA and MAD-Sim dataset with respect to the AUROC metric, explained by dataset composition: the lower AUROC on VisA and MAD-Real stems from the high frequency of ``missing'' component anomalies, which require object-level semantic priors to identify. Our method is not guided by such class semantics, while MultiADS uses class names to gain an advantage on these specific ``absence'' defects. In contrast, the superior AP and F1 scores demonstrate that for visually present anomalies, our defect-aware prompts improve the alignment between the modalities, leading to more accurate localization. More generally, the pattern of a high pixel-level AUROC together with lower AP/F1 reflects the strong class imbalance of MTAS: AUROC rewards correctly ranking the abundant normal background pixels, whereas AP and F1 directly penalize false positives within the predicted defect regions, which favors DAPO's conservative, high-precision predictions that concentrate on the most discriminative abnormal evidence.

\begin{table}[ht]
    \centering
    \caption{Zero-Shot Multi-type Anomaly Segmentation performance. Best results are highlighted in bold. MultiADS results are directly reported from the original paper.}
    \begin{tabular}{c|c|c|ccc}
    \toprule
        \multicolumn{2}{c|}{\textbf{Dataset}} & \multirow{2}{*}{\textbf{Method}} & \multicolumn{3}{c}{\textbf{Pixel-Level}} \\
        \cline{1-2} \cline{4-6}
        Train & Test & & AUROC & AP & F1-score \\
        \hline
        \multirow{8}{*}{MVTec-AD} & \multirow{2}{*}{VisA} & MultiADS & \textbf{93.6} & 24.8 &22.1  \\
        && DAPO (ours) & 88.9 & \textbf{25.0} & \textbf{22.2} \\
        \cline{2-6}
        & \multirow{2}{*}{MPDD} & MultiADS & 95.2 & 53 & 42.8 \\
        && DAPO (ours) & \textbf{95.5} & \textbf{53.6} & \textbf{50.0} \\
         \cline{2-6}\textbf
        & \multirow{2}{*}{MAD-Sim} & MultiADS & \textbf{92.1} & 31.5 & \textbf{27.9} \\
        && DAPO (ours) & 91.9 & \textbf{33.3} & 25.0 \\
         \cline{2-6}
        & \multirow{2}{*}{MAD-Real} & MultiADS & \textbf{89.2} & 52.3 & 52.5 \\
        && DAPO (ours) & 86.1 & \textbf{58.5} & \textbf{58.5} \\
        \bottomrule
    \end{tabular}
    \label{tab:Multi-Type}
\end{table}

\begin{figure}[h]
	\centering
	\begin{subfigure}[b]{\linewidth}
		\centering
		\includegraphics[width=\linewidth]{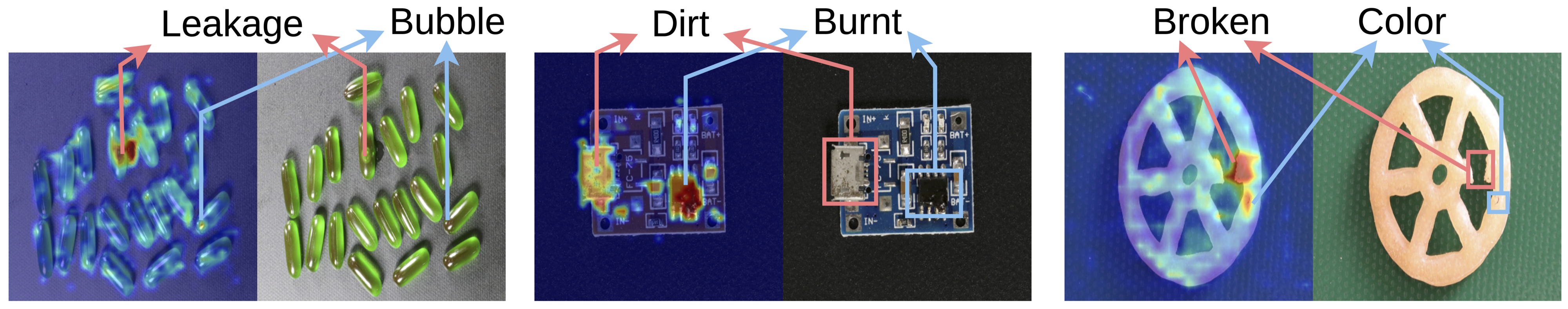}
		\caption{VisA contains multiple defect types (e.g., leakage, bubble, dirt, burnt, broken, and color). We visualize predicted anomaly heatmaps (left) and the corresponding RGB images with defect regions highlighted (right). }
		\label{fig:sub_a}
	\end{subfigure}

	\begin{subfigure}[b]{\linewidth}
		\centering
		\includegraphics[width=\linewidth]{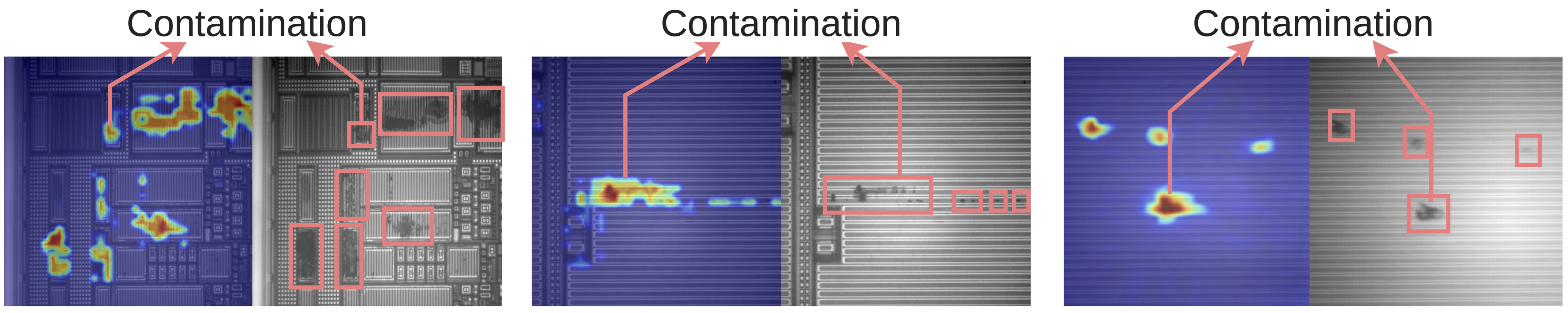}
		\caption{Our internal ASIC dataset focuses on a single defect category (contamination). We show representative examples with predicted anomaly heatmaps (left) and corresponding images (right), where red boxes indicate annotated defect regions. }
		\label{fig:sub_b}
	\end{subfigure}
	\caption{Qualitative examples of DAPO from VisA and our internal ASIC dataset.}
	\label{fig:Qualitative examples}
\end{figure}

To better understand the model's generalization, we analyze individual defect types (Table~\ref{tab:multi_defect_seg}). DAPO generalizes well to unseen(*) defects with a strong, visually-present signal, such as textural or geometric deviations: it reaches 66.6\% F1 on the unseen ``flattening'' defect in MPDD, and on MAD-Real and VisA it identifies ``stains'' (40.4\% F1 vs.\ 6.1\%) and ``breakage'' (20.8\% vs.\ 7.9\%). Averaged over MultiADS across the representative unseen defect types in Table~\ref{tab:multi_defect_seg}, this amounts to a 5.2\% mean improvement in AUROC and F1, underscoring its ability to localize novel defects in the zero-shot setting. The strength fades for anomalies lacking a strong local signal. For fine-grained defects such as ``particle'' on VisA, detection is high (99\% AUROC) but segmentation precision is low: the model recognizes that something is wrong yet cannot localize it precisely. It also struggles with ``missing'' components: the 1.8\% F1 on MAD-Real directly explains the overall AUROC gap in Table~\ref{tab:Multi-Type}, as a missing component leaves a visually normal patch (e.g., the background) that provides no signal for the patch-based ViT encoder to align with a text prompt.

\begin{table}[h]
\centering
\small 
\setlength{\tabcolsep}{3.5pt} 
\caption{Pixel-level segmentation for representative seen and unseen(*) defects across datasets, reported per defect type.}
\label{tab:multi_defect_seg}
\begin{tabular}{ll cc cc} 
\toprule
& & \multicolumn{2}{c}{MultiADS \cite{sadikaj2025multiads}} & \multicolumn{2}{c}{\textbf{DAPO (ours)}} \\
\cmidrule(lr){3-4} \cmidrule(lr){5-6}
\textbf{Dataset} & \textbf{Defect Type} & AUROC & F1 & AUROC & F1 \\
\midrule
\multirow{4}{*}{VisA} 
 & Extra* & 94.1 & 2.1  & \textbf{95.4} & \textbf{3.5} \\
 & Missing* & \textbf{88.5} & 5.3  & 87.6 & \textbf{13.3} \\
 & Particle* & 97.2 & 0.2  & \textbf{99.0} & \textbf{3.1} \\
 & Breakage* & \textbf{98.5} & 7.9  & 97.1 & \textbf{20.8} \\
\midrule
\multirow{4}{*}{MPDD} 
 & Flattening* & 96.7 & 36.1 & \textbf{98.8} & \textbf{66.6} \\
 & Mismatch* & 88.4 & 2.6  & \textbf{93.7} & \textbf{17.9} \\
 & Rust* & 88.4 & 26.1 & \textbf{90.1} & \textbf{28.4} \\
 & Scratch     & 96.7 & \textbf{27.0} & \textbf{97.4} & 26.8 \\
\midrule
MAD-Real 
 & Stains      & 97.0 & 6.1  & \textbf{98.7} & \textbf{40.4} \\
 & Missing     & \textbf{84.1} & \textbf{3.7}  & 77.7 & 1.8 \\
\midrule
MAD-Sim 
 & Burrs* & \textbf{95.6} & 1.2  & 95.4 & \textbf{1.4} \\
 & Stains      & 98.2 & 15.0 & \textbf{99.2} & \textbf{30.0} \\
\bottomrule
\end{tabular}
\end{table}
\subsection{Binary Detection and Segmentation under Distribution Shift} 
\begin{figure}[t]
\centering
\begin{subfigure}[b]{0.49\linewidth}
\centering
\includegraphics[width=\linewidth]{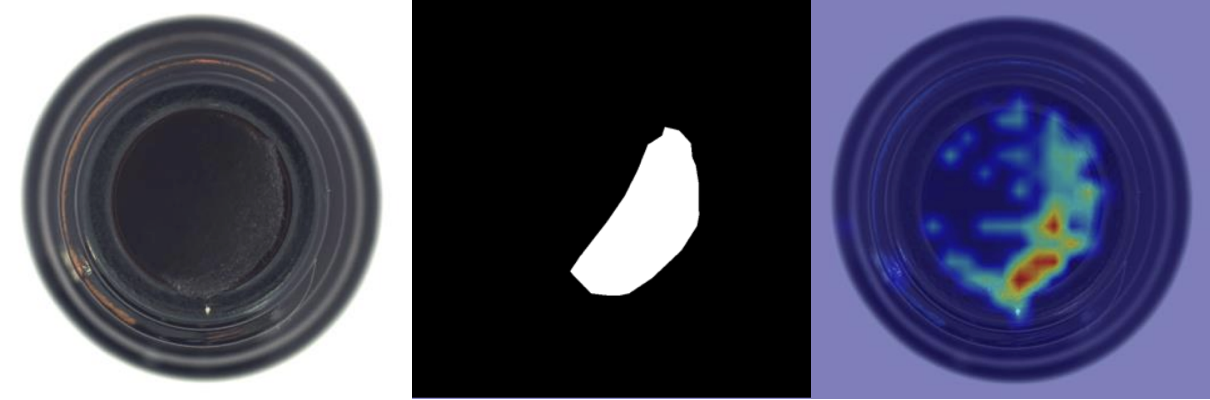}
\caption{Bottle (MVTec-AD)}
\label{fig:large_defect_a}
\end{subfigure}
\hfill
\begin{subfigure}[b]{0.49\linewidth}
\centering
\includegraphics[width=\linewidth]{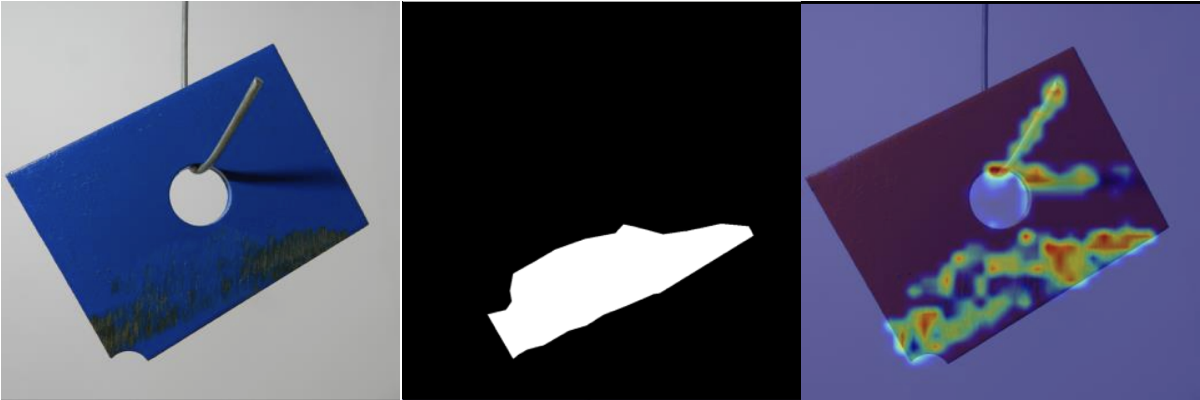}
\caption{Metal Plate (MPDD)}
\label{fig:large_defect_b}
\end{subfigure}
\caption{Visualization of large-area defect predictions. DAPO highlights the most discriminative abnormal evidence rather than expanding to the full annotated region, which lowers AUPRO on broad ground-truth masks while suppressing false positives on normal regions.}
\label{fig:large_defect}
\vspace{-1em}
\end{figure}

Table~\ref{tab:binary_ad} reports binary AD performance under distribution shift at both image and pixel levels. Following prior protocols~\cite{zhou2023anomalyclip, sadikaj2025multiads}, we train on MVTec-AD and evaluate on VisA, MPDD, and Real-IAD as target domains. DAPO achieves the strongest \textbf{image-level} performance across all targets, obtaining the highest AUROC and AP, which indicates robust transfer under substantial domain shift. Averaged over the two strongest defect-aware baselines (AnomalyCLIP and MultiADS) across VisA, MPDD, and Real-IAD, this amounts to a 3.6\% mean improvement in image-level AUROC and AP. For reference, on VisA DAPO attains an image-level AUROC of 84.9, ahead of the recent fine-grained-prompt methods FiLo (83.9) and FAPrompt (84.6) and well above AdaCLIP (75.4) under the same source$\rightarrow$target protocol (FiLo/FAPrompt as reported in their original papers). At the pixel level, our results are competitive, while AUPRO remains below some baselines. Since AUPRO rewards predictions that expand to cover the entire annotated mask, DAPO's conservative heatmaps, which highlight the most discriminative abnormal evidence rather than the full region, are penalized, especially for large or ambiguous defects with broad ground-truth masks (Fig.~\ref{fig:large_defect}). This same conservative behavior is beneficial in deployment: on the internal ASIC dataset, it avoids excessive activation on normal regions and helps reduce false alarms.

\begin{table}[h]
    \centering
    \caption{Binary anomaly detection and anomaly segmentation performance under distribution shift. Best results are in \textbf{bold}. The second-best results are \underline{underlined}.
    }
    \begin{tabular}{c|c|cc|cc}
    \toprule
        \multicolumn{2}{c|}{\textbf{Binary AD and AS}} & \multicolumn{2}{c|}{\textbf{Image-Level}} & \multicolumn{2}{c}{\textbf{Pixel-Level}} \\
        \hline
        Dataset & Method & AUROC & AP & AUROC & AUPRO \\
        \hline
        \multirow{8}{*}{VisA} & CLIP-AC & 65.0 & 70.1 & 24.8  & 17.3  \\
        & CoCoOp & 78.1 & - & 93.6 & - \\
        & WinCLIP & 78.1 & 81.2 & 79.6 & 56.8 \\
        & AprilGAN & 78.0 & 81.4 & 94.2 & 86.8 \\
        & AnomalyCLIP & 82.1 & 85.4 &  \textbf{95.5} & \underline{87.0} \\
        & AdaCLIP & 75.4 & 79.3 & \underline{95.0} & - \\
        & MultiADS & \underline{83.6} & \underline{86.9} & \underline{95.0} &  \textbf{89.7} \\
        \cline{2-6}
        & DAPO (ours) &  \textbf{84.9} &  \textbf{87.1} & 94.3 & 84 \\
        \hline
        \multirow{8}{*}{MPDD} & CLIP-AC & 56.2 & 66.0 & 58.7 & 29.1   \\
        & CoCoOp & 61 & - & 95.1 & - \\
        & WinCLIP & 63.6 & 69.9 & 76.4 & 48.9  \\
        & AprilGAN & 73.0 & 80.2 & 94.1 & 83.2  \\
        & AnomalyCLIP & 77.0 & 82.0 &  \textbf{96.5} & 88.7  \\
        & AdaCLIP & 66.3 & 75 & \underline{96.3} & - \\
        & MultiADS & 78.3 & 78.4 & 95.8 &  \textbf{89.7} \\
        \cline{2-6}
        & DAPO (ours) &  \textbf{81.2} &  \textbf{83.6}  & 95.1 & 84.4 \\
        \hline
        \multirow{6}{*}{Real-IAD} & WinCLIP & 75 & 72.3 & 87.1 & 59.9   \\
        & AprilGAN & 75.7 & 73.5 & 96 & 86.8   \\
        & AnomalyCLIP & 78.4 & 76.7 & 96.2 & 85.7  \\
        & AdaCLIP & 70.1 & 68.5 & 95.3 & - \\
        & MultiADS & 78.7 & 79.1 &  \textbf{96.6} &  \textbf{87.1}   \\
        \cline{2-6}
        & DAPO (ours) & \textbf{84.3} &  \textbf{84}  & \underline{96.4} & 80.3 \\
        \bottomrule
    \end{tabular}
    \label{tab:binary_ad}
    \vspace{-1.5em}
\end{table}
Importantly, we additionally evaluate on an internal ASIC dataset (Table~\ref{tab:binary_ad_internal}), which reflects a more deployment-realistic setting than public benchmarks. On this dataset, AnomalyCLIP exhibits a marked performance degradation (68.2 AUROC / 35.3 AP), whereas defect-aware approaches (MultiADS and DAPO) remain substantially more stable. This gap suggests that explicitly incorporating defect semantics can improve robustness when moving from curated benchmarks to real production data, consistent with the qualitative examples in Fig.~\ref{fig:Qualitative examples}.

\begin{figure}[ht]
  \centering
  \begin{subfigure}[b]{0.33\textwidth}
    \includegraphics[width=\linewidth]{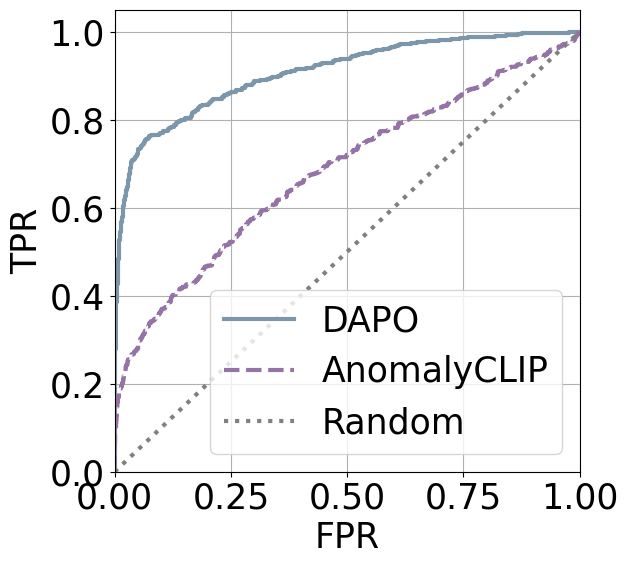}
    \caption{ROC}
    \label{fig:roc}
  \end{subfigure}
  \hfill
  \begin{subfigure}[b]{0.3\textwidth}
    \includegraphics[width=\linewidth]{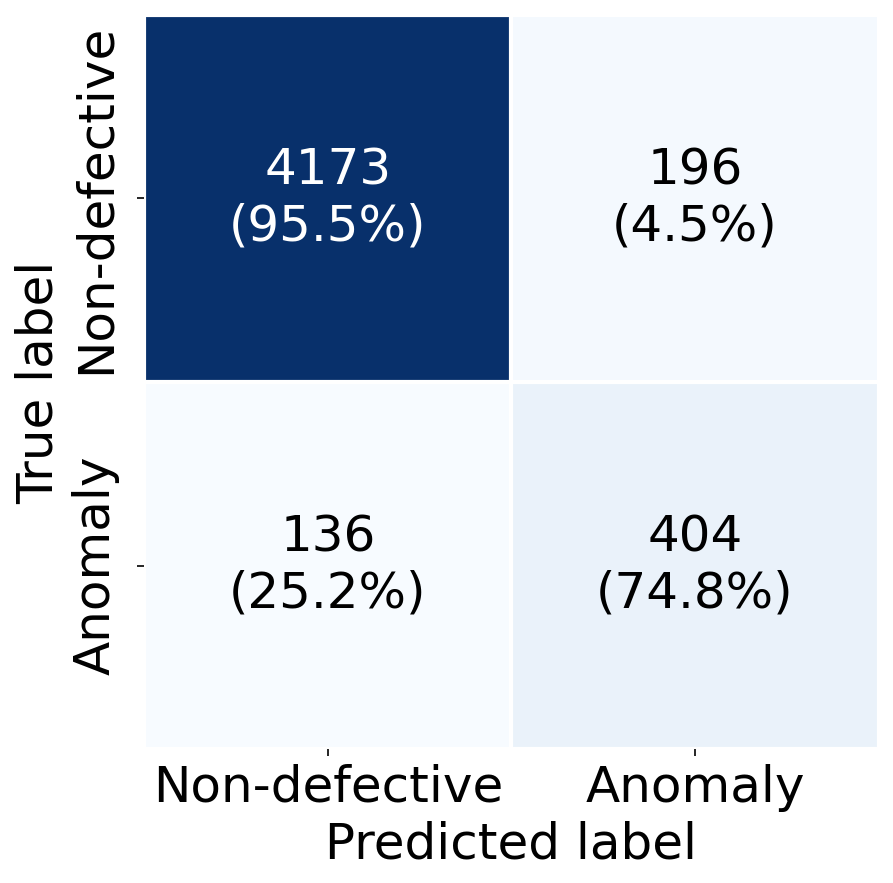}
    \caption{DAPO}
    \label{fig:confusion matrix1}
  \end{subfigure}
  \hfill
  \begin{subfigure}[b]{0.3\textwidth}
    \includegraphics[width=\linewidth]{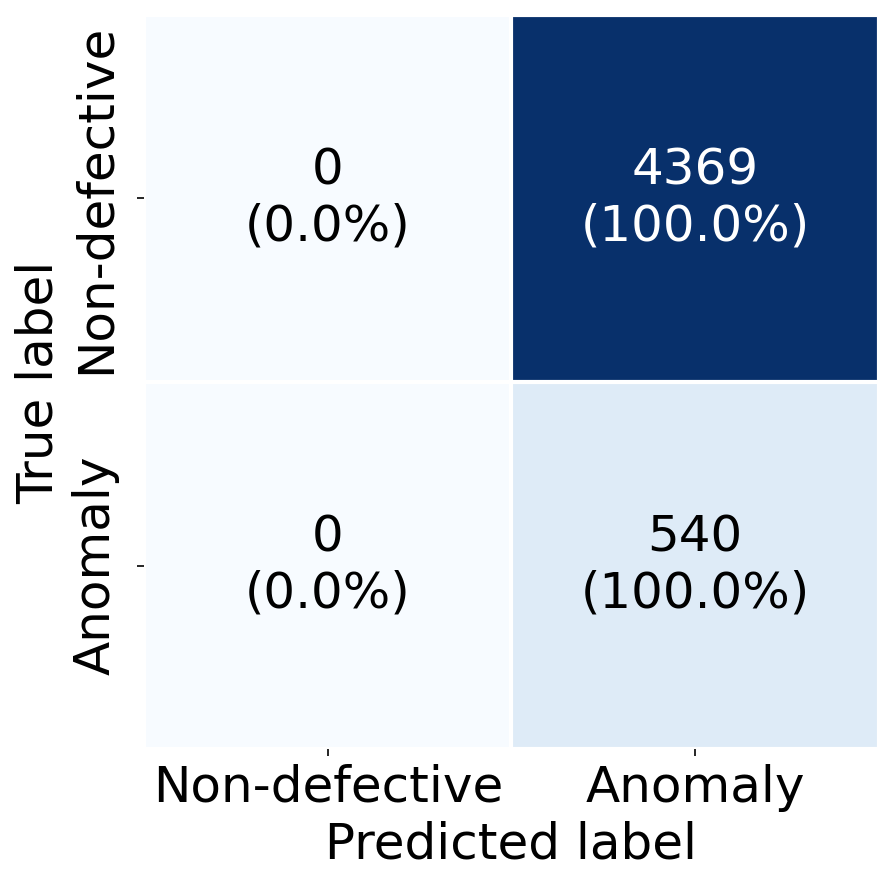}
    \caption{AnomalyCLIP}
    \label{fig:confusion matrix2}
  \end{subfigure}
  \caption{Binary anomaly detection performance comparison. (a) ROC curves for DAPO, AnomalyCLIP, and Random Classifier. (b) and (c) are confusion matrices, demonstrating classification accuracy, with our approach DAPO.}
  \label{fig:combined}
\end{figure}

Fig.~\ref{fig:combined} further analyzes behavior on the internal dataset. The ROC curves (Fig.~\ref{fig:roc}) show that DAPO consistently dominates AnomalyCLIP across operating points, indicating better separability between normal and contaminated samples under production conditions. At a fixed threshold, the confusion matrices (Fig.~\ref{fig:confusion matrix1},~\ref{fig:confusion matrix2}) reveal a balanced trade-off for DAPO (TN=4173, TP=404), operating at FPR=4.5\% and TPR=74.8\%, with low false alarms and strong defect recall, an important property for deployment. In contrast, AnomalyCLIP assigns the anomaly label to nearly all samples, yielding an impractically high false-alarm rate.

\begin{table}[htb]
    \centering
    \caption{Binary anomaly detection performance under distribution shift on our internal semiconductor dataset.}
    \begin{tabular}{c|c|cc}
    \toprule
        \multicolumn{2}{c|}{\textbf{Binary Anomaly Detection}} & \multicolumn{2}{c}{\textbf{Image-Level}} \\
        \hline
        Dataset & Method & AUROC & AP \\
        \hline
        \multirow{3}{*}{\makecell{Semi-conductor \\ (Internal Dataset)}} & AnomalyCLIP & 68.2 & 35.3  \\
        & MultiADS & 91.1 & 61.7 \\
        \cline{2-4}
        & DAPO (ours) &  \textbf{91.8} &  \textbf{79.1} \\
        \bottomrule
    \end{tabular}
    \label{tab:binary_ad_internal}
\end{table}



\subsection{Ablation Study}  
We further conduct a prompt decomposition study to isolate the contribution of each component in DAPO: (a) fixed defect-name anchors (\textit{text-only}), (b) shared learnable context tokens without defect anchors (\textit{soft-only}), and (c) their hybrid composition (DAPO). As shown in Table~\ref{tab:prompt_ablation_defs}, text-only prompts preserve semantic specificity but remain limited by the image–text modality gap, leading to weak transfer in practice. In contrast, soft-only prompts substantially improve binary anomaly detection by learning task-adaptive context, yet they lack explicit type cues and therefore fail to produce well-separated decision regions required for $(K{+}1)$-channel multi-type anomaly segmentation (MTAS). Combining shared context tokens with defect-name anchors yields the best of both worlds: the anchors provide explicit defect semantics, while the learnable tokens capture transferable anomaly context and improve cross-domain alignment, resulting in the strongest AUROC/AP and enabling MTAS.
The lower block of Table~\ref{tab:prompt_ablation_defs} isolates three further design choices. Removing token sharing and learning a separate prompt per defect type (\emph{per-defect tokens}) drops AUROC to 70.9, confirming that a shared, type-agnostic context, rather than memorizing individual defect prompts, is what enables transfer; this directly validates our central design choice. Disabling the progressive injection of learnable tokens (\emph{w/o progressive fine-tuning}) reduces AUROC to 79.7, and replacing the lightweight linear adapters with deeper 2-layer adapters brings no gain (80.5). Together these confirm that DAPO's improvements stem from its compositional, shared-token design rather than from additional trainable capacity.

\begin{table}[H]
\centering
\footnotesize
\caption{Component ablations on MPDD (image-level AUROC/AP). The upper block isolates the prompt composition (defect-name anchor vs.\ shared learnable tokens); the lower block ablates token sharing, progressive fine-tuning, and adapter depth. \cmark/\xmark~indicate whether the variant supports the $(K{+}1)$-channel MTAS task.}
\label{tab:prompt_ablation_defs}
\setlength{\tabcolsep}{6pt}
\begin{tabular}{l c c c}
\toprule
Variant & AUROC & AP & MTAS \\
\midrule
Defect-anchor only (text, non-learnable) & 53.8 & 65.0 & \cmark \\
Soft-only (no defect anchor) & 73.7 & 78.1 & \xmark \\
\midrule
Per-defect tokens (no sharing) & 70.9 & 75.6 & \cmark \\
w/o progressive fine-tuning & 79.7 & 81.5 & \cmark \\
2-layer adapters & 80.5 & 82.3 & \cmark \\
\midrule
\textbf{DAPO (shared tokens + anchor)} & \textbf{81.2} & \textbf{83.6} & \cmark \\
\bottomrule
\end{tabular}
\end{table}

%% file: sec/7_conclusion.tex
\section{Conclusion}
We presented DAPO, a defect-aware prompt optimization framework for binary anomaly detection/segmentation and zero-shot multi-type anomaly segmentation. DAPO learns compositional hybrid prompts that combine a small set of shared learnable context tokens with explicit defect-descriptor anchors, enabling prompt instantiation for new defect descriptors at inference. By aligning global and patch-level visual representations with defect-aware text semantics, it improves fine-grained localization and robustness under distribution shift, achieving consistent gains in image-level detection and competitive segmentation across five public benchmarks and an internal dataset. Future work includes extending DAPO to other vision-language backbones and improving localization of absence-based structures via object-level priors and few-shot adaptation.

\section*{Acknowledgements}
The authors thank the International Max Planck Research School for Intelligent Systems (IMPRS-IS)
for supporting Hongkuan Zhou.

%% file: sec/X_suppl.tex
\noindent
In the appendix, we discuss the key statistics of the five publicly available datasets and one internal dataset. Additionally, the baselines we compared are briefly described. Finally, additional experiments with detailed results, along with visualizations, are listed.  
\section{Datasets}
In this section, we describe the public benchmark industrial datasets, MvTec-AD~\cite{bergmann2019mvtec}, VisA~\cite{zou2022spot}, MPDD~\cite{jezek2021deep}, MAD~\cite{zhou2023pad}, Real-IAD~\cite{wang2024real}, along with an internal dataset.  Table~\ref{tab:datasets} summarizes key statistics of these datasets, including the number of distinct product classes and the distribution of normal and anomalous samples. Each product may exhibit multiple defect types, which are categorized into their respective defect classes as introduced in MultiADS~\cite{sadikaj2025multiads}. More detailed defect-level descriptions for each dataset can be found in the supplementary material of the original MultiADS paper. 
Our internal dataset consists of high-resolution semiconductor chip images collected from an industrial environment, featuring a single known defect type. It provides realistic, real-world conditions for evaluating the robustness of the proposed system under distribution shift. For our internal dataset, only bounding box annotations were available for the segmentation task, which were used to generate corresponding anomaly masks. Hence, the pixel-level scores are not accurate for evaluating the localization performance. The details of the defect types are mentioned in the Table~\ref{tab:datasets}.

\begin{table}[ht]
\fontsize{8pt}{10pt}\selectfont
\centering{
\caption{Key statistics of the datasets used for industrial anomaly detection.}
\label{tab:datasets}
\renewcommand{\arraystretch}{1}
\setlength{\tabcolsep}{4pt} 
\begin{tabular}{l|l|l|c|c|l}
\toprule
\textbf{Dataset} &  $|\mathcal{C}|$ & \textbf{\begin{tabular}[c]{@{}c@{}}\#Normal /\\ \#Anomalous\end{tabular}} & \textbf{Usage} \\
\midrule
MVTec AD  & 15 & (467; 1258) & Industrial defect detection \\
VisA  &  12 & (962; 1200) & Industrial defect detection \\
MPDD & 6 & (176; 282) & Industrial defect detection \\
MAD & 20 & (5,231; 4902) & Simulated defect detection \\
Real-IAD & 30 & (99,721; 51,326) & Industrial defect detection \\
\midrule
\begin{tabular}[c]{@{}l@{}}Semi-conductor\\(internal dataset)\end{tabular} & 1 & (4,369; 540) & Industrial defect detection \\
\bottomrule
\end{tabular}
}
\end{table}

\begin{table}[ht]
\fontsize{7pt}{10pt}\selectfont
\setlength{\tabcolsep}{4pt}
    \centering
    \caption{Detailed information on the defects of our internal dataset}
    \begin{tabular}{|c|c|c|c|c|}
    \hline
      \multirow{2}{*}{\textbf{Object}}   & \multirow{2}{*}{\textbf{Defects}} & \multirow{2}{*}{\textbf{Defect Type}} & \multicolumn{2}{c|}{Test Data (No.s)}  \\
      \cline{4-5}
      & & & \textbf{Normal} & \textbf{Anomaly} \\
      \hline
      \makecell{Semi-Conductor \\ chip}  &  Contamination & Contamination & 4369 & 540 \\
      \hline
    \end{tabular}
    \label{tab:internal-dataset}
\end{table}

\section{Baselines}
To evaluate the performance of our DAPO, we compare DAPO with several baseline models. The results of the baselines are taken directly from the respective papers.  Details of the respective baselines are given as follows:
\begin{itemize}
    \item \textbf{CLIP-AC}~\cite{radford2021learning}: They use an ensemble of text prompts designed for normal and abnormal classes, e.g., \textit{a photo of a normal [cls]}, and \textit{a photo of a damaged [cls]}, where \textit{[cls]} denotes the target class name. The generated embeddings are averaged to represent the normal and abnormal classes respectively. The anomaly scores are computed using the contrastive equation from~\cite{radford2021learning}. Results are taken from~\cite{sadikaj2025multiads}. For anomaly segmentation, the similarity score checking is extended to local embeddings extracted from patch features.
    \item  \textbf{WinClip}~\cite{jeong2023winclip}: WinClip is the primary baseline for anomaly detection under distribution shift. It uses a set of handcrafted prompts specific to the anomaly detection domain. For segmentation, a window-based approach is used due to the inherent limitation of pre-trained CLIP. The results are directly taken from the~\cite{sadikaj2025multiads} paper.
    \item \textbf{AprilGAN~\cite{chen2023april}}: This approach extends the WinClip by introducing an additional linear adapter to achieve better accurate segmentation results. The results are taken directly from the~\cite{sadikaj2025multiads}.
    \item  \textbf{CoCoOp~\cite{zhou2022conditional}}: CoCoOp extend the CoOp~\cite{zhou2022learning}. To adapt CoOp for anomaly detection, authors of~\cite{zhou2023anomalyclip} extends the learnable text prompt templates $[V_1][V_2]\dots[V_n][\text{cls}]$ with $[V_1][V_2]\dots[V_n][\text{anomaly}][\text{cls}]$. Due to the sensitivity of the class distribution shift, CocoOp builds a light neural network to generate an instance conditional embeddings that adapts the learned prompts to the respective test set domains. The results are directly taken from the paper~\cite{sadikaj2025multiads}.
    \item  \textbf{AnomalyCLIP~\cite{zhou2023anomalyclip}}: AnomalyCLIP is the primary prompt learning baseline for anomaly detection. They learn object-agnostic prompts that capture anomaly semantics. The learnable tokens are structured similarly to CoOp, but instead of including [cls] information, they use object-level semantics: $[V_1][V_2]\dots[V_n][\text{anomaly}][\text{object}]$, where [object] refers to the object as whole.  The results are directly taken from their original paper~\cite{zhou2023anomalyclip}.
    \item  \textbf{AdaCLIP~\cite{cao2024adaclip}}: AdaCLIP uses a hybrid prompt strategy combining static templates and dynamically generated prompts per test instance for CLIP adaptation. This hybrid design enables anomaly detection under distribution shifts. The results for AdaCLIP are directly taken from their original paper.
    \item  \textbf{MultiADS~\cite{sadikaj2025multiads}}: MultiADS is the key baseline for multi-defect anomaly segmentation in zero-shot, and anomaly detection under distribution shifts . They use extensive prompt templates associated with each defect type and a product class. For e.g., \textit{a photo of a [cls] having a [D] defect.}. Here the \textit{[cls]}, and \textit{[D]} are replaced with class and defect information. The embeddings are then compared with the patch-level scores to generate local anomaly maps. The anomaly scores are computed by aggregating known anomaly scores. The results are directly from their paper.
\end{itemize}

\section{Experiments}
Here, we will discuss in detail on DAPO through the experiments, and the ablation we carried out. We will also display the visualization and fine-grained results of our approach.

\subsection{Implementation Details}
As mentioned in the ``implementation details'' subsection of main paper, we use \textbf{VIT-L/14@336} model pretrained on \textbf{OpenAI}. The model is optimized using Adam~\cite{kingma2014adam}, with learning rate $0.001$, and a batch size of $8$. The number of intermediate patch-level features, $M$ is 4, specifically from layers $6, 12, 18, 24$. In line with prior works, we also follow a transfer learning strategy: training on Mvtec-AD and evaluating on other datasets, and train on VisA to test on Mvtec-AD. Input images are resized to $518$ and normalized following the CLIP preprocessing pipeline. All our experiments are conducted for 5 epochs, with Pytorch-2.7.0, on a single NVIDIA-H200 GPU. Empirically, we found observed that performance peaks around epoch 1 or 2, with later epochs exhibiting overfitting to the training dataset. 

\subsection{Initialization of Prompts}
\begin{figure}[t]
\centering
\begin{tikzpicture}
\begin{axis}[
    ybar,
    width=\linewidth,
    height=6cm,
    bar width=7pt,
    enlarge x limits=0.25,
    ylabel={AUROC (\%)},
    ymin=70,
    symbolic x coords={Random, CLIPSpace, Offset},
    xtick=data,
    legend style={at={(0.5,-0.25)}, anchor=north, legend columns=2},
    legend image code/.code={
        \draw[#1, draw=none] (0cm,-0.1cm) rectangle (0.3cm,0.1cm);
    },
    xticklabel style={font=\footnotesize},
]

\addplot+[fill=light-purple, draw=none] coordinates {
    (Random,72.2)
    (CLIPSpace,84.9)
    (Offset,85)
};
\addlegendentry{VisA (Image)}

\addplot+[fill=dark-purple, draw=none] coordinates {
    (Random,93.3)
    (CLIPSpace,94.3)
    (Offset,91.7)
};
\addlegendentry{VisA (Pixel)}

\addplot+[fill=light-blue, draw=none] coordinates {
    (Random,74.8)
    (CLIPSpace,81.2)
    (Offset,80.3)
};
\addlegendentry{MPDD (Image)}

\addplot+[fill=dark-blue, draw=none] coordinates {
    (Random,95.6)
    (CLIPSpace,95.1)
    (Offset,93.4)
};
\addlegendentry{MPDD (Pixel)}

\end{axis}
\end{tikzpicture}
\caption{Performance comparison of token initialization strategies: \textbf{Random} ($\mathcal{N}(0,1)$), \textbf{CLIP-Space} (aligned to CLIP's $\mu, \sigma$), and \textbf{Offset} (abnormal tokens shifted by $5\sigma$). CLIP-Space provides faster convergence and higher pixel-level accuracy.}
\label{fig:initialization_comparison}
\end{figure}
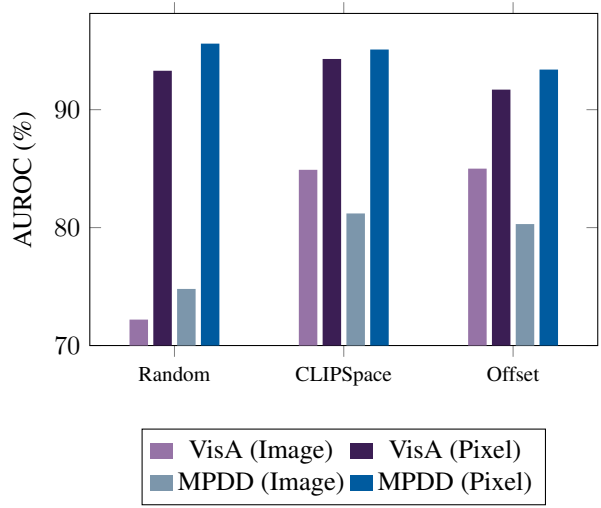

We study the initialization strategy of the learnable tokens: normal context tokens $V$, abnormal context tokens $W$ and layer tokens. We conducted experiments using three distinct initialization strategies: We used random initialization and found that the model achieved its best performance only in later epochs, suggesting the model required a longer duration to stabilize and find the objective minimum. Secondly, we initialized the context and layer tokens by aligning them to the mean ($\mu$) and standard deviation ($\sigma$) of the CLIP's token embeddings. We found that when initialized within CLIP's feature space, the model converged much faster and yielded better overall results compared to the random strategy. Finally, in order to explicitly separate the feature space of normal and abnormal context tokens, we shifted the initialization of abnormal tokens by an offset 5$\sigma$ from the mean of the CLIP space. We and found that this required more epochs to achieve its best image-level results. However, the pixel level accuracy over these epochs was lower compared with the non-offset approach. In Figure~\ref{fig:initialization_comparison}, we present the Image and Pixel level results on the VisA and MPDD for each initialization strategy, using the best performing epoch achieved across 5 training cycles.  

\subsection{Hyperparameter Study}
We investigate the sensitivity of DAPO to the local-loss weight $\lambda$ and the prompt length $l$. Fig.\ref{fig:ablation}(a) shows that increasing $\lambda$ from small values initially improves performance, indicating that stronger patch-level (multi-type) supervision helps align defect-aware semantics and benefits detection under distribution shift. Fig.\ref{fig:ablation}(b) further shows that longer prompts tend to hurt performance. We attribute this to the increased number of trainable prompt parameters, which makes optimization more prone to overfitting on the source training data and reduces cross-domain generalization.
\begin{figure}[h]
    \centering
    \begin{subfigure}[b]{0.49\columnwidth}
        \centering
        \begin{subfigure}[t]{0.49\columnwidth}
            \centering
            \begin{tikzpicture}
            	\begin{axis}[
            		width=1.3\linewidth,
            		height=1.0\linewidth,
            		xlabel={},
            		ylabel={AUROC \%},
            		label style={font=\fontsize{7}{8}\selectfont},
            		tick label style={font=\fontsize{7}{8}\selectfont},
            		ylabel style={font=\fontsize{7}{8}\selectfont, yshift=-5pt},
            		grid=major,
            		axis background/.style={fill=gray!10},
            		grid style={line width=0.1pt, draw=white},
            		axis on top=false,
            		axis line style={draw=none},
            		tick style={draw=none},
            		every axis plot/.append style={line width=0.6pt, mark size=2.0pt},
                    legend style={
                    	at={(0.5,1.02)},        
                    	anchor=south,           
                    	font=\fontsize{6}{7}\selectfont,
                    	legend columns=2,       
                    	draw=none, fill=none,
                    	inner xsep=0pt, inner ysep=0pt,
                    	nodes={inner sep=0pt, outer sep=0pt},
                    },
                    legend cell align=left,
                    legend image post style={scale=0.8},
            	]
                \addplot+[color=light-blue, mark=*, mark options={fill=light-blue}] coordinates {
                    (1, 93.9) (2, 93.0) (3, 94.4) (4, 95.1)
                };
                \addlegendentry{MPDD}
                
                \addplot+[color=light-purple, mark=square*, mark options={fill=light-purple}] coordinates {
                    (1, 90.3) (2, 92.6) (3, 91.7) (4, 94.3)
                };
                \addlegendentry{VisA}
                \end{axis}
            \end{tikzpicture}
        \end{subfigure}
        \begin{subfigure}[t]{0.49\columnwidth}
            \centering
            \begin{tikzpicture}
            	\begin{axis}[
            		width=1.3\linewidth,
            		height=1.0\linewidth,
            		xlabel={},
            		ylabel={AUPRO \%},
            		label style={font=\fontsize{7}{8}\selectfont},
            		tick label style={font=\fontsize{7}{8}\selectfont},
            		ylabel style={font=\fontsize{7}{8}\selectfont, yshift=-5pt},
            		grid=major,
            		axis background/.style={fill=gray!10},
            		grid style={line width=0.1pt, draw=white},
            		axis on top=false,
            		axis line style={draw=none},
            		tick style={draw=none},
            		every axis plot/.append style={line width=0.6pt, mark size=2.0pt},
                    legend style={
                    	at={(0.5,1.02)},        
                    	anchor=south,           
                    	font=\fontsize{6}{7}\selectfont,
                    	legend columns=2,       
                    	draw=none, fill=none,
                    	inner xsep=0pt, inner ysep=0pt,
                    	nodes={inner sep=0pt, outer sep=0pt},
                    },
                    legend cell align=left,
                    legend image post style={scale=0.8},
            	]
            \addplot+[color=light-blue, mark=*, mark options={fill=light-blue}] coordinates {
                (1, 84.2) (2, 83.8) (3, 82.4) (4, 84.4)
            };
            \addlegendentry{MPDD}
            
            \addplot+[color=light-purple, mark=square*, mark options={fill=light-purple}] coordinates {
                (1, 80.6) (2, 81.3) (3, 81.0) (4, 84.0)
            };
            \addlegendentry{VisA}
            \end{axis}
            \end{tikzpicture}
        \end{subfigure}
        \caption{ Ablation on the weight term $\lambda$}
        \label{subfig:a}
    \end{subfigure}
    \hfill
    \begin{subfigure}[b]{0.49\columnwidth}
        \centering
        \begin{subfigure}[t]{0.49\columnwidth}
            \centering
            \begin{tikzpicture}
            \begin{axis}[
                xtick={5, 6,7,8,9,10},
                width=1.3\linewidth,
                height=1.0\linewidth,
                xlabel={},
                ylabel={AUROC \%},
                label style={font=\fontsize{7}{8}\selectfont},
                tick label style={font=\fontsize{7}{8}\selectfont},
                ylabel style={font=\fontsize{7}{8}\selectfont, yshift=-5pt},
                grid=major,
                axis background/.style={fill=gray!10},
                grid style={line width=0.1pt, draw=white},
                axis on top=false,
                axis line style={draw=none},
                tick style={draw=none},
                every axis plot/.append style={line width=0.6pt, mark size=2.0pt},
                legend style={
                    at={(0.5,1.02)},        
                    anchor=south,           
                    font=\fontsize{6}{7}\selectfont,
                    legend columns=2,       
                    draw=none, fill=none,
                    inner xsep=0pt, inner ysep=0pt,
                    nodes={inner sep=0pt, outer sep=0pt},
                },
                legend cell align=left,
                legend image post style={scale=0.8},
            ]
            \addplot+[color=light-blue, mark=*, mark options={fill=light-blue}] coordinates {
               (5, 80.6) (6, 78.7) (7, 76.2) (8, 75.0) (9, 74.9) (10, 74.7)
            };
            \addlegendentry{MPDD}
            
            \addplot+[color=light-purple, mark=square*, mark options={fill=light-purple}] coordinates {
                 (5, 85.6) (6, 84.2) (7, 84.9) (8, 85.3) (9, 84.9) (10, 85.1)
            };
            \addlegendentry{VisA}
            \end{axis}
            \end{tikzpicture}
        \end{subfigure}
        \hfill
        \begin{subfigure}[t]{0.49\columnwidth}
            \centering
            \begin{tikzpicture}
            \begin{axis}[
                xtick={5, 6,7,8,9,10},
                width=1.3\linewidth,
                height=1.0\linewidth,
                xlabel={},
                ylabel={AP \%},
                label style={font=\fontsize{7}{8}\selectfont},
                tick label style={font=\fontsize{7}{8}\selectfont},
                ylabel style={font=\fontsize{7}{8}\selectfont, yshift=-5pt},
                grid=major,
                axis background/.style={fill=gray!10},
                grid style={line width=0.1pt, draw=white},
                axis on top=false,
                axis line style={draw=none},
                tick style={draw=none},
                every axis plot/.append style={line width=0.6pt, mark size=2.0pt},
                legend style={
                    at={(0.5,1.02)},        
                    anchor=south,           
                    font=\fontsize{6}{7}\selectfont,
                    legend columns=2,       
                    draw=none, fill=none,
                    inner xsep=0pt, inner ysep=0pt,
                    nodes={inner sep=0pt, outer sep=0pt},
                },
                legend cell align=left,
                legend image post style={scale=0.8},
            ]
            \addplot+[color=light-blue, mark=*, mark options={fill=light-blue}] coordinates {
                (5, 83.6) (6, 82.3) (7, 81.4) (8, 80.1) (9, 79.7) (10, 78)
            };
            \addlegendentry{MPDD}
            
            \addplot+[color=light-purple, mark=square*, mark options={fill=light-purple}] coordinates {
                (5, 88.3) (6, 87.8) (7, 87.9) (8, 87.3) (9, 87.1) (10, 87.8)
            };
            \addlegendentry{VisA}
            \end{axis}
            \end{tikzpicture}
        \end{subfigure}
        \caption{Ablation on learnable prompt length $l$}
        \label{subfig:b}
    \end{subfigure}
    \caption{Ablation study on the validation split. We vary (a) the local-loss weight $\lambda$ in Eq.~\ref{eq:loss} and (b) the learnable prompt length $l$, while keeping all other settings fixed.}
    \label{fig:ablation}
\end{figure}

\subsection{Prototype Aggregation}
To determine the robust representation of the final anomaly prototype, $z^{D_k}$, We perform ablation on the aggregation method used to combine information from multiple defect prompts. Baseline: By default, we define the anomaly prototype prompt by calculating the simple average across all $K$ defect-type prompts. This treated all defects equally, which can be noisy if only a few prompts are relevant to a specific image. Attention-weighted averaging: To filter out redundant or irrelevant prompts, we experimented with a weighted average based on the relevance of the prompt on the image. We calculated the attention weight $w_i$ associated with each defect prompt embedding and the image embedding. The final prototype was then calculated as the weighted sum: $\frac{1}{K}\sum_{j=1}^{K}w_j*\mathbf{z}^{\text{D}_k}$. We found that using an attention-weighted prototype did not yield significant performance improvement over the simple baseline mean. Both image, and pixel level scores showed only marginal gains or decrease in performance across the dataset, Table~\ref {tab:prototype_ablation}. 

\begin{table}[t]
\centering
\caption{Prototype Aggregation Ablation: Simple Mean vs. Attention Weighting. The table display the Pixel and Image level AUROC scores on VisA and MPDD}
\label{tab:prototype_ablation}
\begin{tabular}{l c c c c}
    \toprule
    \textbf{Method} & \textbf{Dataset} & \textbf{Image($\%$)} & \textbf{Pixel ($\%$)} & \textbf{$\Delta$ AP} \\
    \midrule
    Baseline & VisA & \textbf{84.9} & \textbf{94.3} & $+3\%$ \\
    Attention & VisA & 79.9 & 93.0 & $-3\%$\\
    \midrule
    Baseline & MPDD & \textbf{81.2} & \textbf{95.1} & $+3.2\%$ \\
    Attention & MPDD & 78.0 & 91.9 & $-3.2\%$ \\
    \bottomrule
    \multicolumn{5}{l}{\footnotesize $\Delta$ AP is the average change relative to the baseline for that dataset.}
\end{tabular}
\end{table}

\subsection{Prefix tuning approach}
In this section, we demonstrate the effectiveness of the progressive tuning strategy compared to baseline prefix tuning. We train our model in MVTec-AD and evaluate it in MPDD. MPDD is selected for this evaluation due to its challenging conditions, including variable spatial orientations, diverse object positions and distances from the camera. The image-level AUROC and AP scores using progressive connection and traditional prefix tuning across different depths are shown in Figure~\ref{fig:progressive-vs-non}. Detailed results of the best-performing configuration are provided in Table~\ref{tab:mpdd_img_level}. We observe that with progressive connection, the model consistently achieves over 80\% AP on MPDD, with minor fluctuations in AUROC, and clearly outperforms the traditional approach. These results highlight the importance of gradually adapting the encoders to better align image and text features, thereby improving anomaly detection performance across datasets. Here we keep the length of the learnable prompts, $l$ as 5, and the number of learnable tokens as 4.
\begingroup
\setlength{\abovecaptionskip}{2pt}
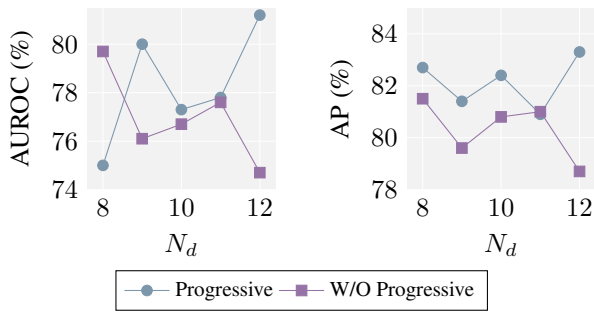
\begin{figure}[t]
\centering
\begin{subfigure}[t]{0.49\columnwidth}
  \centering
  \vspace{0pt}
  \begin{tikzpicture}
    \begin{axis}[
      legend to name=NdLegend,
      xlabel={$N_d$},
      ylabel={AUROC (\%)},
      ymin=74, ymax=81.5,
      grid=major,
      axis background/.style={fill=gray!10},
      grid style={line width=0.1pt, draw=white},
      axis on top=false,
      axis line style={draw=none},
      tick style={draw=none},
      width=\linewidth,
      height=4.0cm,
      legend style={
        at={(0.5, -0.45)},
        anchor=north,
        font=\footnotesize,
        legend columns=2},
    ]
      \addplot+[color=light-blue, mark=*, mark options={fill=light-blue}] coordinates {
        (8, 75.0) (9, 80.0) (10, 77.3) (11, 77.8) (12, 81.2)
      };
      \addlegendentry{Progressive}

      \addplot+[color=light-purple, mark=square*, mark options={fill=light-purple}] coordinates {
        (8, 79.7) (9, 76.1) (10, 76.7) (11, 77.6) (12, 74.7)
      };
      \addlegendentry{W/O Progressive}
    \end{axis}
  \end{tikzpicture}
\end{subfigure}
\hfill
\begin{subfigure}[t]{0.49\columnwidth}
  \centering
  \vspace{0pt}
  \begin{tikzpicture}
    \begin{axis}[
      xlabel={$N_d$},
      ylabel={AP (\%)},
      ymin=78, ymax=85,
      grid=major,
      axis background/.style={fill=gray!10},
      grid style={line width=0.1pt, draw=white},
      axis on top=false,
      axis line style={draw=none},
      tick style={draw=none},
      width=\linewidth,             
      height=4.0cm,                 
    ]
      \addplot+[color=light-blue, mark=*, mark options={fill=light-blue}] coordinates {
        (8, 82.7) (9, 81.4) (10, 82.4) (11, 80.9) (12, 83.3)
      };

      \addplot+[color=light-purple, mark=square*, mark options={fill=light-purple}] coordinates {
        (8, 81.5) (9, 79.6) (10, 80.8) (11, 81) (12, 78.7)
      };
    \end{axis}
  \end{tikzpicture}
\end{subfigure}
\begin{center}
\vspace{-0.7em} 
  \pgfplotslegendfromname{NdLegend}
\end{center}

\caption{Ablation on the AUROC and AP with different $N_d$ on MPDD using progressive and naive prefix tuning.}
\label{fig:progressive-vs-non}
\end{figure}
\endgroup
\subsection{Prefix Depth}
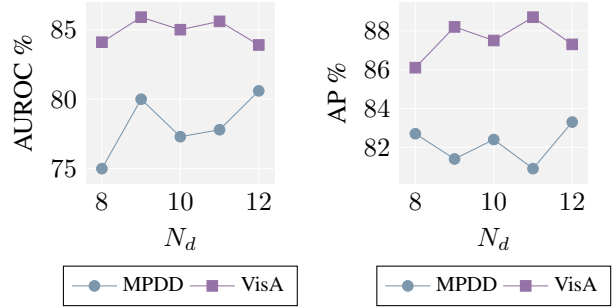
\begin{figure}[t]
    \centering
    
    \begin{subfigure}[t]{\columnwidth}
        \centering
        \begin{subfigure}[t]{0.49\columnwidth}
            \centering
            \begin{tikzpicture}
            \begin{axis}[
                xlabel={$N_d$},
                ylabel={AUROC \%},
                grid=major,
                axis background/.style={fill=gray!10},
                grid style={line width=0.1pt, draw=white},
                axis on top=false,
                axis line style={draw=none},
                tick style={draw=none},
                legend style={at={(0.5,-0.45)}, anchor=north, font=\footnotesize, legend columns=2},
                legend cell align={left},
                width=\textwidth,
                height=4.0cm
            ]
            \addplot+[color=light-blue, mark=*, mark options={fill=light-blue}] coordinates {
                (8, 75.0) (9, 80.0) (10, 77.3) (11, 77.8) (12, 80.6)
            };
            \addlegendentry{MPDD}
            
            \addplot+[color=light-purple, mark=square*, mark options={fill=light-purple}] coordinates {
                (8, 84.1) (9, 85.9) (10, 85.0) (11, 85.6) (12, 83.9)
            };
            \addlegendentry{VisA}
            \end{axis}
            \end{tikzpicture}
        \end{subfigure}
        \hfill
        \begin{subfigure}[t]{0.49\columnwidth}
            \centering
            \begin{tikzpicture}
            \begin{axis}[
                xlabel={$N_d$},
                ylabel={AP \%},
                grid=major,
                axis background/.style={fill=gray!10},
                grid style={line width=0.1pt, draw=white},
                axis on top=false,
                axis line style={draw=none},
                tick style={draw=none},
                legend style={at={(0.5,-0.45)}, anchor=north, font=\footnotesize, legend columns=2},
                legend cell align={left},
                width=\textwidth,
                height=4.0cm
            ]
            \addplot+[color=light-blue, mark=*, mark options={fill=light-blue}] coordinates {
                (8, 82.7) (9, 81.4) (10, 82.4) (11, 80.9) (12, 83.3)
            };
            \addlegendentry{MPDD}
            
            \addplot+[color=light-purple, mark=square*, mark options={fill=light-purple}] coordinates {
                (8, 86.1) (9, 88.2) (10, 87.5) (11, 88.7) (12, 87.3)
            };
            \addlegendentry{VisA}
            \end{axis}
            \end{tikzpicture}
        \end{subfigure}
        \label{subfig:prefix}
    \end{subfigure}
    \caption{Effect on injecting learnable tokens at different depth of the text encoder.}
    \label{fig:prefix_depth}
\end{figure}

We study the impact of learnable tokens injection depth $N_d$ on detection performance under distribution shift. The result shows that the relationship between $N_d$ and detection performance is non-monotonic. Specifically, we observed an initial gain in performance followed by a small dip at intermediate depth, before reaching the maximum depth, Figure~\ref{fig:prefix_depth}. This non-linear trend indicated that maximal integration is required for stability, as intermediate layers are highly sensitive. Therefore, for the final configuration, we utilized the full tested depth to have consistent performance. Please note that this ablation was performed only on the text encoder side of the architecture, as the visual prompt tuning was held constant at maximum layer depth (fully integrated into all visual transformer layers).

\subsection{Analysis of Feature Space}
\begin{figure}[t]
    \centering
    \begin{subfigure}[b]{0.49\columnwidth}
        \centering
        \includegraphics[width=\linewidth]{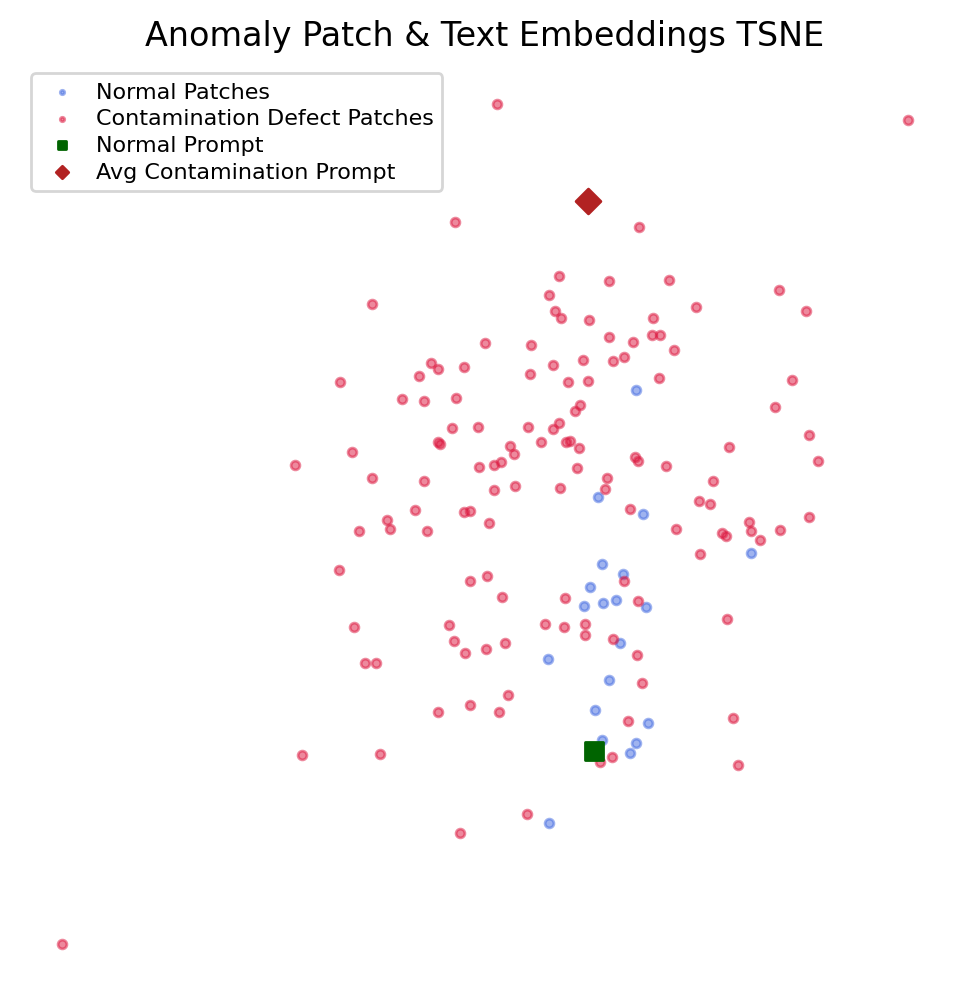} 
        \caption{DAPO (2 Images)}
        \label{fig:sparse}
    \end{subfigure}
    \hfill 
    \begin{subfigure}[b]{0.49\columnwidth}
        \centering
        \includegraphics[width=\linewidth]{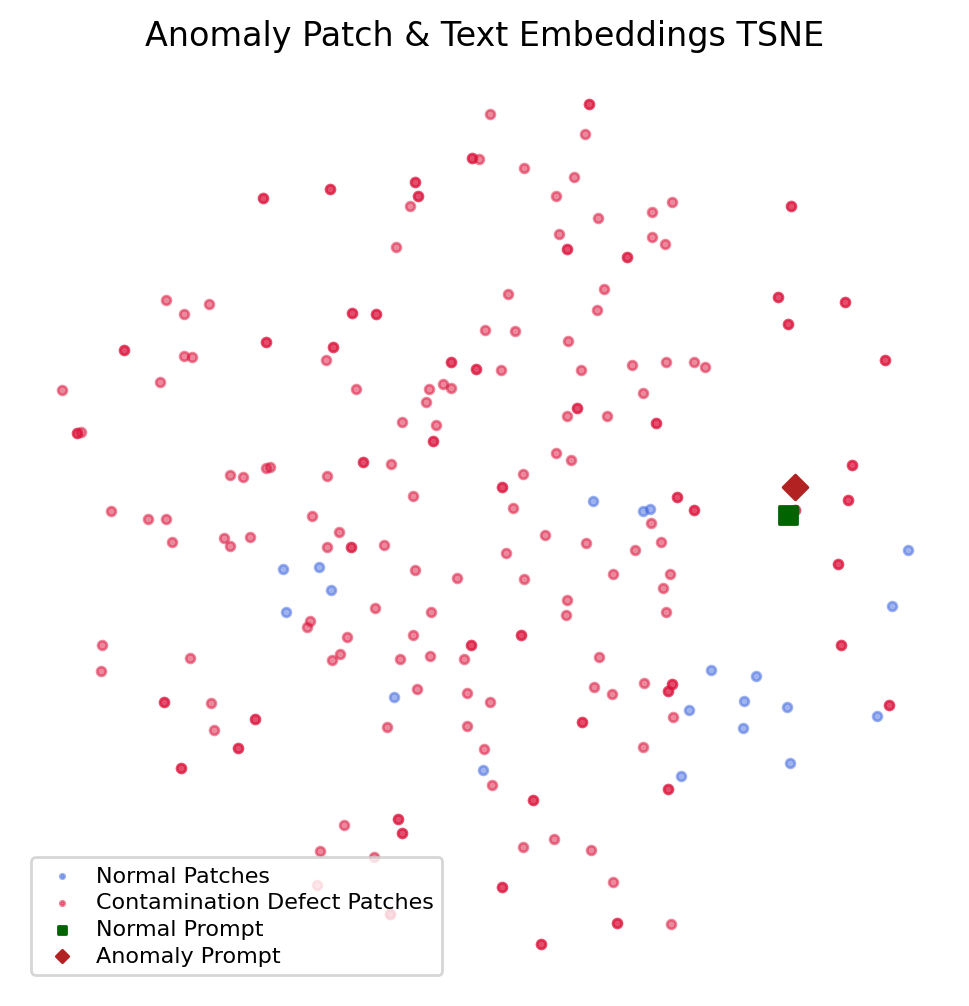}
        \caption{AnomalyCLIP (2 Images)}
        \label{fig:dense}
    \end{subfigure}
    
    \caption{Comparison of feature topology. (a) Shows how patches are distributed and closer with the text-embedding in DAPO. (b) Shows how patches are scattered, with text embeddings in close proximity to each other in AnomalyCLIP.}
    \label{fig:tsne_comparison}
\end{figure}

To visually verify the fine-grained feature topology of the learned embeddings, we utilized t-SNE~\cite{maaten2008visualizing} to project the high dimensional feature space into a 2D space. Our visualization comprises two distinct projections: a sample of 2 anomaly images (Figure~\ref{fig:tsne_comparison} left) from DAPO, and the same sample of anomaly images (Figure~\ref{fig:tsne_comparison} right) from AnomalyCLIP. In the projection, we observe a clear Euclidean separation between the normal and defect text anchor (``contamination" defect-type) in DAPO, as well as reasonable separation between anomaly (red) and normal (blue) patches, where normal patches are much closer towards the normal prompt. It shows that for individual samples, the model projects the normal and anomaly features into regions, creating a high local margin that facilitates better classification on instances. However, as in AnomalyCLIP, the topology shifts. The projection is scattered and the text anchors move into close proximity. This proximity treats defects as subtle perturbations of the normal classes, rather than as a separate disjoint domain. This could be the reason we believe that AnomalyCLIP was classifying every image as an anomaly in our internal dataset.

\subsection{Additional Results}
We demonstrate our model’s fine-grained, data-subset-level performance under distribution shifts. In Tables~\ref{tab:binary_visa_mpdd}, we present the image-level and pixel-level results of our method, DAPO, evaluated on individual subsets of MPDD and VisA. These results are reported for anomaly detection under distribution shifts, a setting commonly referred to in the literature as zero-shot anomaly detection. Tables~\ref{tab:binary_mad_combined} present the results of our approach on the simulated datasets. We observe that our model consistently achieves the best or second-best performance at the image level, indicating its effectiveness in defect detection. It is important to note that both datasets contain numerous missing component anomalies, which contribute to the overall lower performance, particularly for MAD-Sim, which includes a significantly higher number of product images compared to MAD-Real. Table~\ref{tab:realiad} shows the performance of our model on the Real-IAD dataset, a recent benchmark that contains approximately 20,000 test samples. Our method consistently achieves the best performance at the image level, demonstrating high AUROC and AP, even when compared to few-shot approaches such as MultiADS~\cite{sadikaj2025multiads} and AdaCLIP~\cite{cao2024adaclip}, our method achieves near best performance as reported in the supplementary section of the MultiADS paper.

\begin{table*}[t]
\centering
\caption{Comparison of image-level performance on MPDD between progressive connection (DAPO) and without progressive connection, with overall mean per method.}
\label{tab:mpdd_img_level}
\setlength{\tabcolsep}{4pt}
\renewcommand{\arraystretch}{1.3}
\resizebox{\textwidth}{!}{
\begin{tabular}{l|ccc|ccc|ccc|ccc|ccc|ccc|ccc}
\toprule
\textbf{Method} 
& \multicolumn{3}{c|}{\textbf{Bracket Black}} 
& \multicolumn{3}{c|}{\textbf{Bracket Brown}} 
& \multicolumn{3}{c|}{\textbf{Bracket White}} 
& \multicolumn{3}{c|}{\textbf{Connector}} 
& \multicolumn{3}{c|}{\textbf{Metal Plate}} 
& \multicolumn{3}{c|}{\textbf{Tubes}} 
& \multicolumn{3}{c}{\textbf{Mean}} \\
\cline{2-22}
 & AUROC & AP & F1 & AUROC & AP & F1 & AUROC & AP & F1 & AUROC & AP & F1 & AUROC & AP & F1 & AUROC & AP & F1 & AUROC & AP & F1 \\
\midrule
W/O Progressive & 70.3 & 75.7 & 81.8 & 63.2 & 81.5 & 81.0 & 77.9 & 72.9 & 80.0 & 76.9 & 63.0 & 66.7 & 92.1 & 97.2 & 92.6 & 97.5 & 98.8 & 95.0 & 79.7 & 81.5 & \textbf{82.8} \\
DAPO            & 73.6 & 78.9 & 81.5 & 59.6 & 77.3 & 80.6 & 85.4 & 83.1 & 81.7 & 75.2 & 65.0 & 63.4 & 96.1 & 98.7 & 93.5 & 97.0 & 98.8 & 95.5 & \textbf{81.2} & \textbf{83.6} & 82.7 \\
\bottomrule
\end{tabular}
}
\end{table*}

\subsection{Internal Dataset}
\begin{figure}[t]
    \centering
    \includegraphics[width=0.7\textwidth]{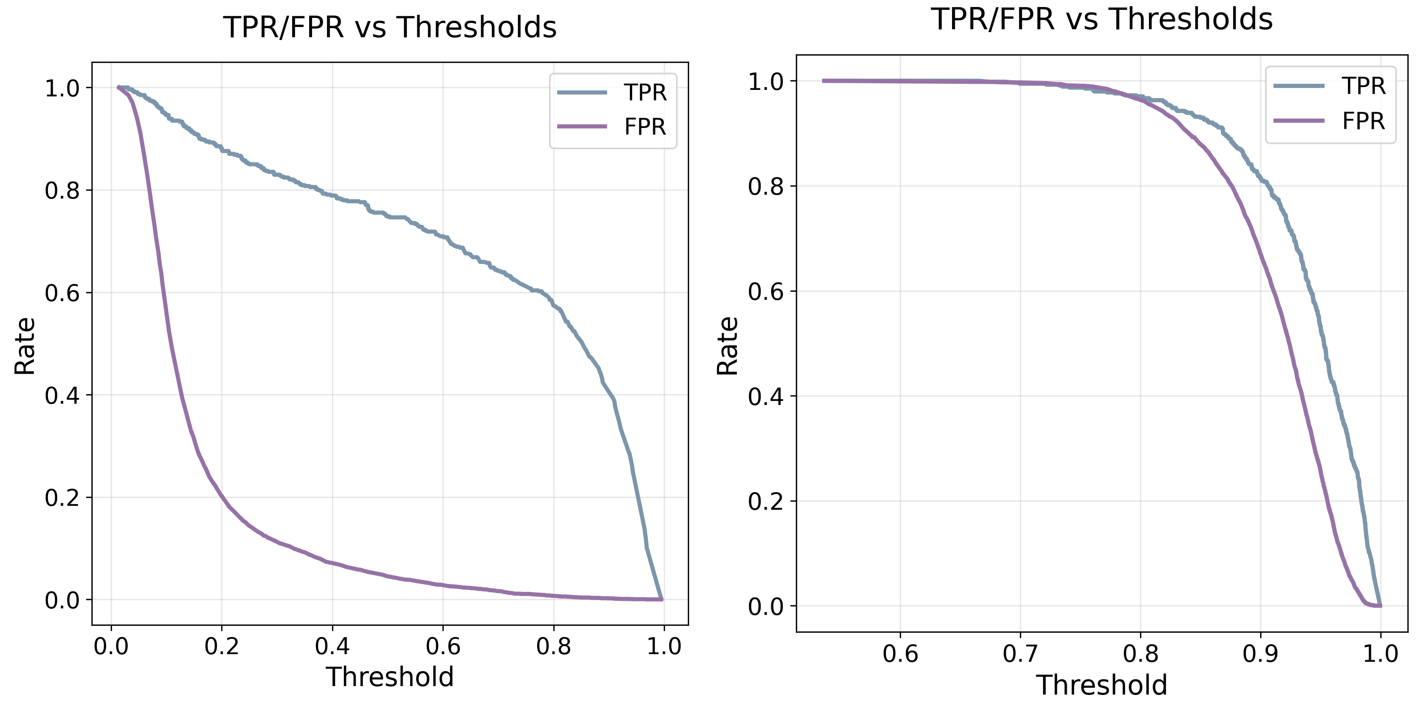}
    \caption{TPR and FPR across different threshold for anomaly detection under distribution shift. DAPO (ours) on Left, AnomalyClip on Right}
    \label{fig:tprvsfpr}
\end{figure}

To evaluate our model's robustness across varying thresholds in detecting anomalies on our internal semiconductor chip dataset, we plotted the False Positive Rate (FPR) and True Positive Rate (TPR) against different thresholds (see Figure~\ref{fig:tprvsfpr}). As illustrated, our model consistently outperforms the baseline AnomalyCLIP across all thresholds, demonstrating a stronger separation between the anomaly and normality spaces, and highlighting the discriminative power of our approach.

The consistent performance gap between our approach and AnomalyCLIP across all threshold values indicates the robustness of our progressive connection mechanism in handling distribution shifts inherent in real manufacturing environments. This stability is crucial for practical applications where threshold selection must balance detection sensitivity with operational efficiency. Furthermore, the smooth ROC curves demonstrate the reliability of our anomaly scoring mechanism, suggesting effective feature representations that capture anomaly patterns while maintaining discriminative power against normal variations in the manufacturing process.

\subsection{Exploratory Study: Multi-label defect classification}
While the primary focus of the work is binary anomaly detection and multi-type/binary anomaly segmentation, we extended our framework to explore \textbf{multi-label defect classification}. In this setting, we aim to identify the specific defect types present in an image (e.g., distinguishing between a \textit{scratch}, \textit{crack}, or \textit{contamination}). Unlike standard multi-class classification which utilizes a Softmax distribution, real-world industrial defects are often non-exclusive. To address this, we treated the problem as a multi-label task. For a given image features, we computed the similarity logits with all potential defect text prompts and the normal prompt. To obtain independent probabilities for each defect type, we utilized a relative scoring mechanism. We subtracted the logit of the normal class from each defect class logit to center the scores around the decision boundary, followed by a Sigmoid activation:
\begin{equation}
    P(d_i) = \sigma(\text{Sim}(z^x, z^{D_k}) - \text{Sim}(z^x, z^{N}))
\end{equation}
where $\sigma$ is the sigmoid function. A defect $D_k$ is predicted if $P(D_k) > 0.5$.
We evaluated the Precision, Recall, and F1-Score (Macro-averaged) on the VisA and MPDD datasets. The results are presented in Table~\ref{tab:exploratory_visa_mpdd}. The model exhibits high Recall ($>90\%$ in many categories), indicating that defective images consistently trigger the defect prompts. Conversely, the Precision scores are consistently low ($<30\%$). This indicates a high rate of false positive predictions for specific defect classes. These results suggest that while the learnable tokens are effective at distinguishing \textit{anomalous} samples from \textit{normal} ones (driving the high Recall), they do not form mutually exclusive clusters for fine-grained defect types. The feature representations of similar defects remain entangled in the latent space, limiting the model's ability to perform precise multi-type classification at the image level. However, for the downstream task of segmentation, this over-activation ensures that the pixel-level anomaly maps capture the defect region, regardless of the specific label ambiguity. 

\subsection{Multi Type Anomaly Segmentation} 
We demonstrate our model’s ability to localize specific defect types, rather than merely detecting anomalies. In particular, we introduce a zero-shot setting, where defect types marked with (*) are unseen during training. This setting encourages future research on more generalized and robust defect segmentation. Table~\ref{tab:mad_multitype},~\ref{tab:mpdd_multitype},~\ref{tab:visa_multitype} present our model's performance on the MAD (Simulated and Real), MPDD and VisA datasets, respectively. These results highlight the model’s effectiveness in segmenting both seen and unseen defect types. 

\section{Visualization}
\begin{figure}[H]
\centering
\includegraphics[width=\textwidth]{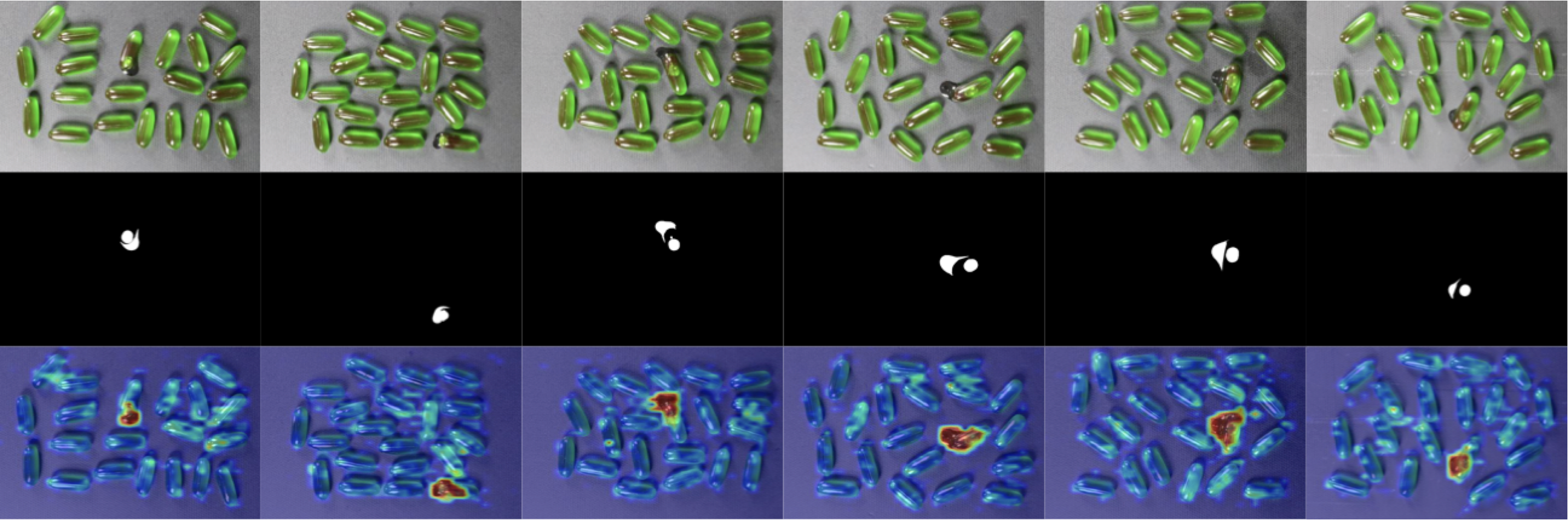}
\caption{The visualization highlights the \textbf{capsule} object from the VisA dataset, while the model is trained on the MVTec-AD dataset. The first row represents the input, the second row shows the ground truth segmentation mask. The last row presents the segmentation results from DAPO. Despite the distribution shift, our model accurately localizes both \textbf{leak} and \textbf{bubble} defects. Notably, the image contains multiple defects, demonstrating the model's capability to segment multiple anomalies within a single sample. }
\label{fig:visa_capsule}
\end{figure}

\begin{figure}[h]
\centering
\includegraphics[width=\textwidth]{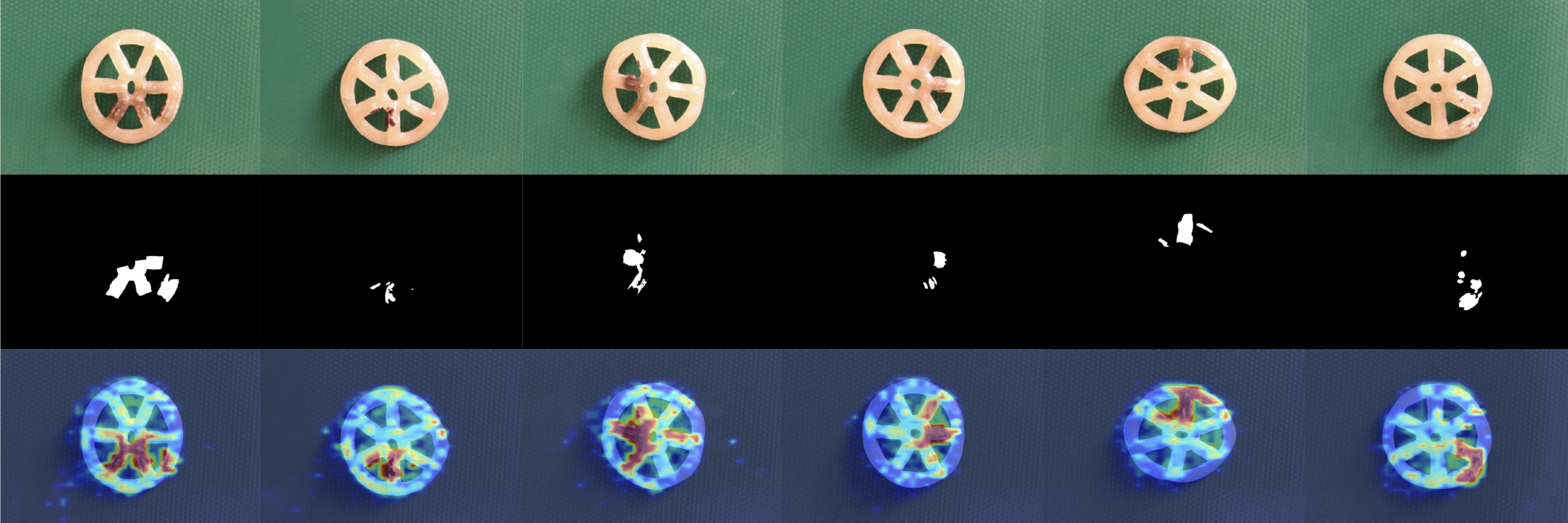}
\caption{The visualization highlights the \textbf{Fryum} object from the VisA dataset, with the model trained on MVTec-AD. As shown, DAPO successfully localizes the \textbf{burnt} defect, however the segmentation slightly overextends beyond the actual defect region, indicating minor over-segmentation. }
\label{fig:visa_fryum}
\end{figure}

\begin{figure}[h]
\centering
\includegraphics[width=\textwidth]{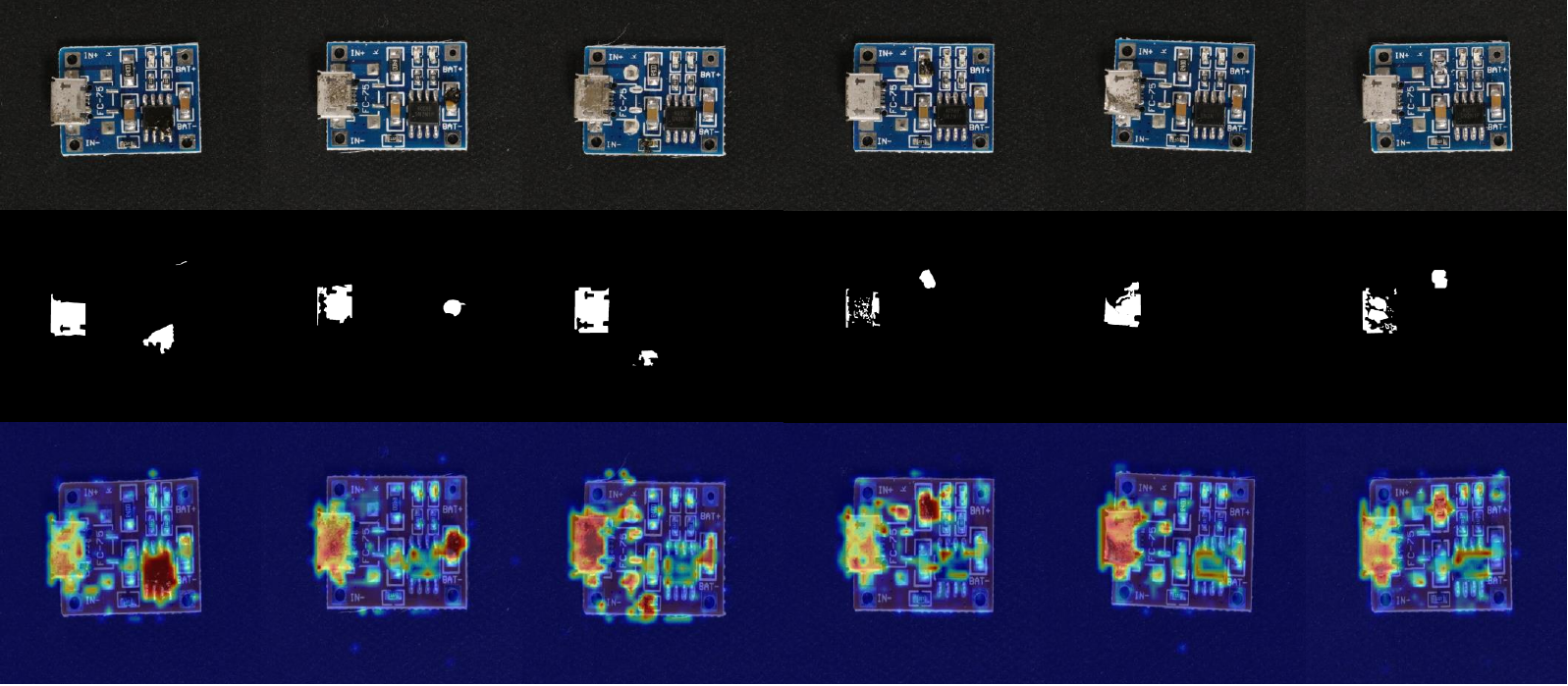}
\caption{The visualization highlights the \textbf{PCB4} object from the VisA dataset, with the model trained on MVTec-AD. As shown, DAPO successfully localizes both the \textbf{burnt} and \textbf{dirt} defects within the same image. However, it fails to detect a small \textbf{scratch} defect in the top-right corner of the first image, likely due to the patch-level features being unable to effectively capture such fine-grained anomalies.}
\label{fig:visa_pcb}
\end{figure}

\begin{figure}[h]
\centering
\includegraphics[width=\textwidth]{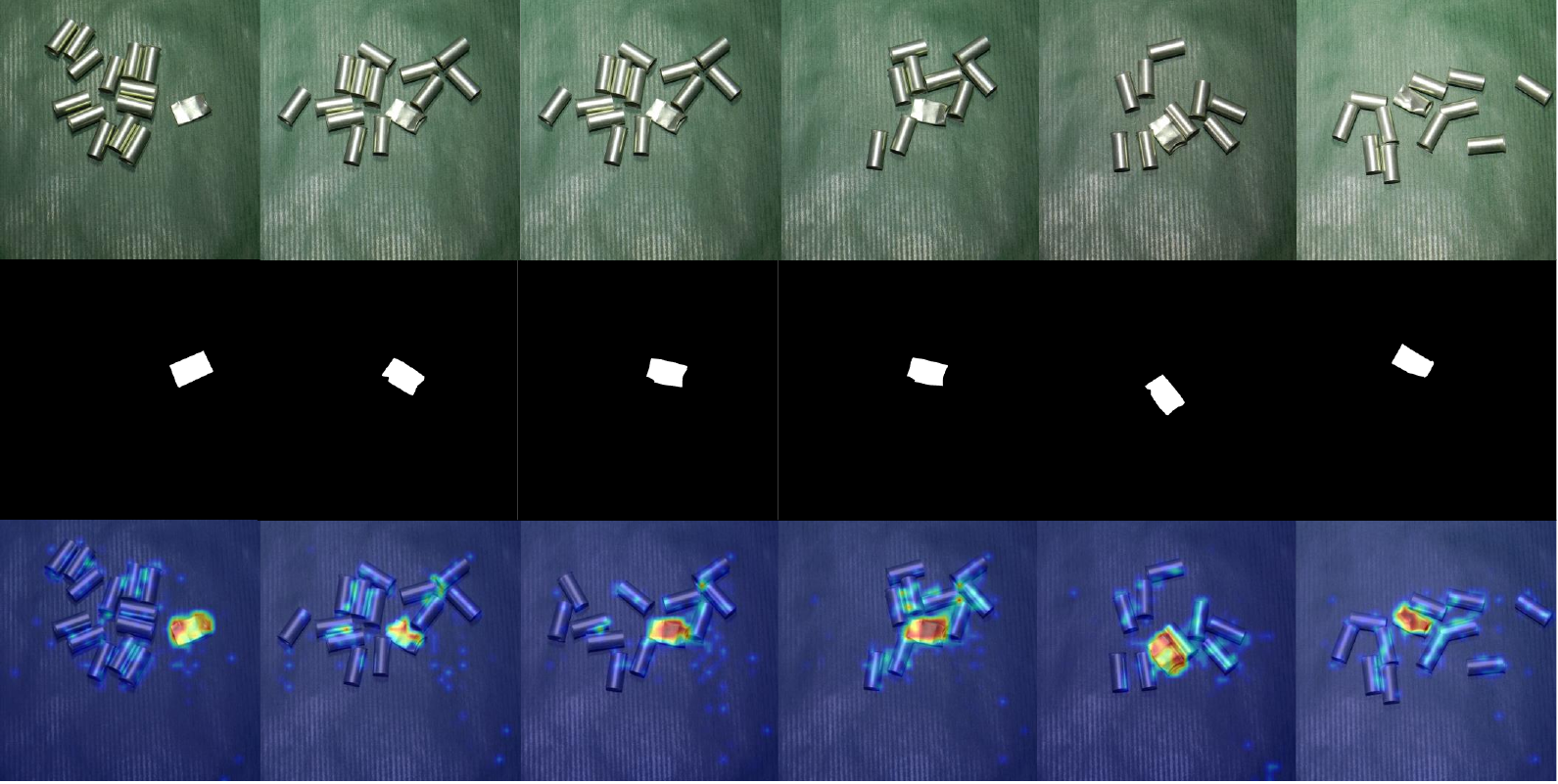}
\caption{The visualization highlights the \textbf{Tubes} object from the MPDD dataset. The model, trained on MVTec-AD, accurately localizes the \textbf{flattening} defect with minimal noise, demonstrating its precision in segmenting flattening defects, as evident from Table~\ref{tab:mpdd_multitype}.}
\label{fig:mpdd_tubes}
\end{figure}

\begin{figure}[h]
\centering
\includegraphics[width=\textwidth]{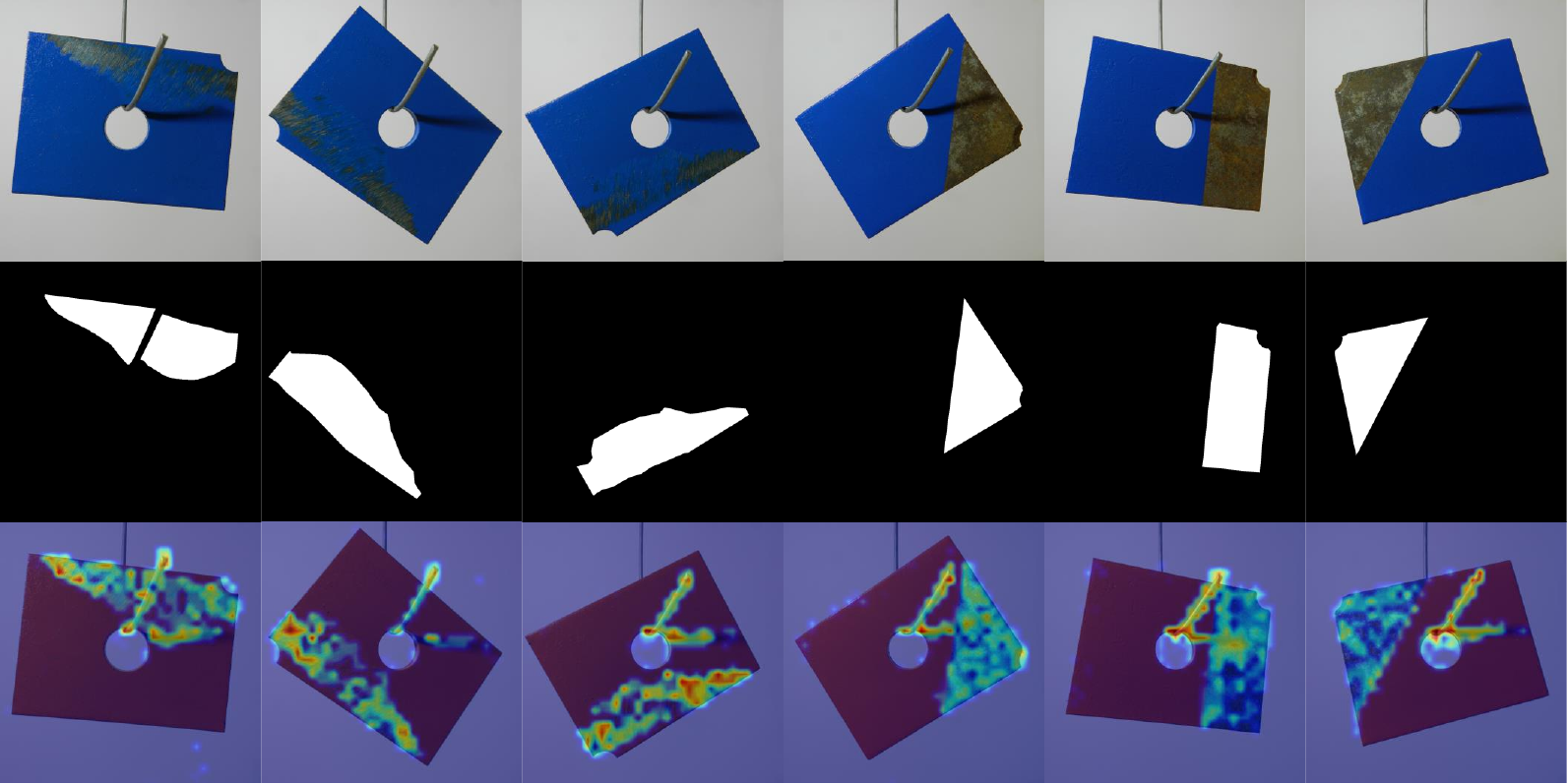}
\caption{The visualization highlights the \textbf{Metal Plate} object from the MPDD dataset. Columns 1–3 show accurate localization of the \textbf{scratch} defect, while columns 4–6 demonstrate precise segmentation of the \textbf{rust} defect.}
\label{fig:mpdd_metalplate}
\end{figure}

\begin{figure}[h]
\centering
\includegraphics[width=\textwidth]{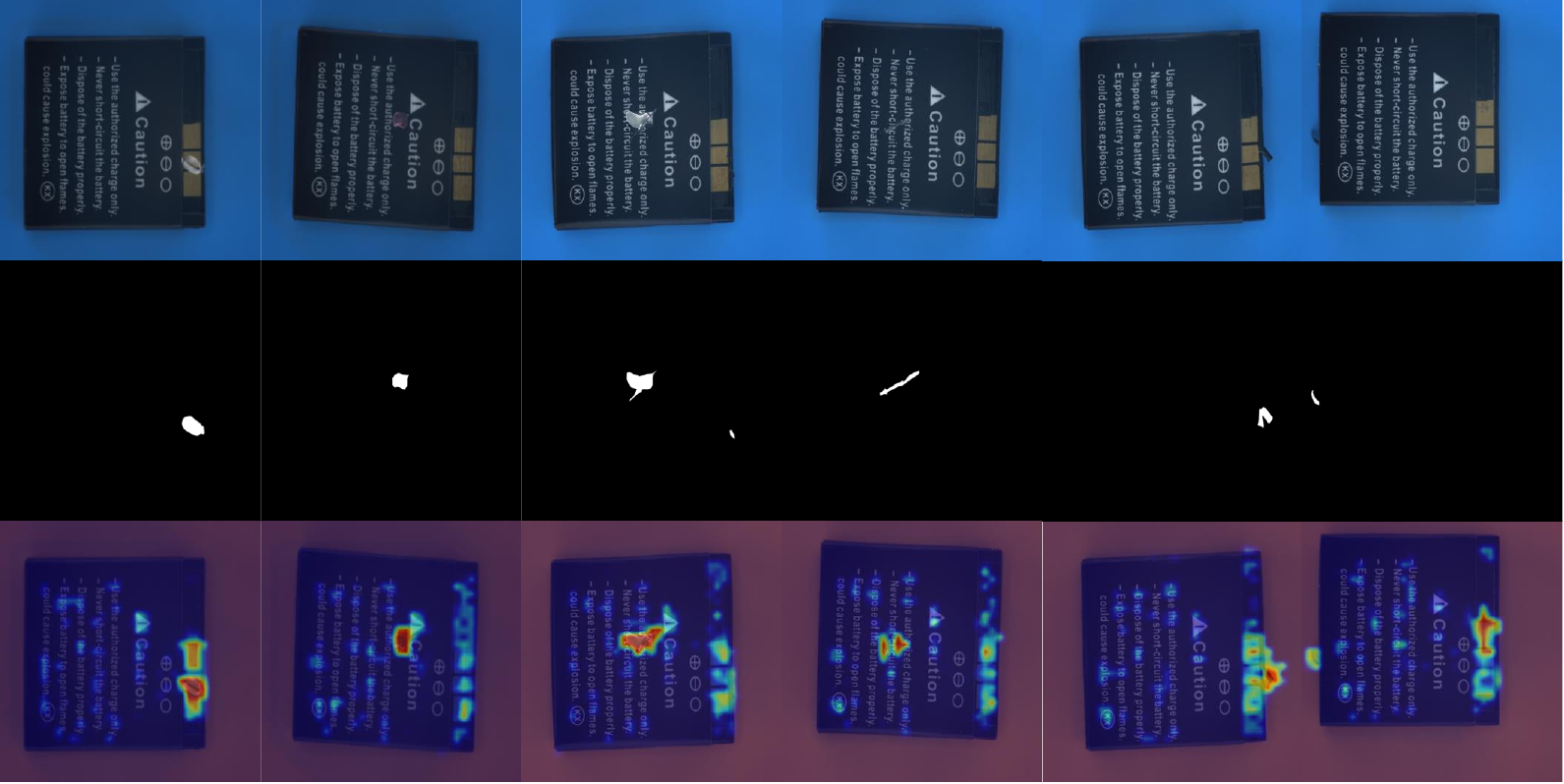}
\caption{The visualization shows the \textbf{Phone Battery} object from the Real-IAD dataset. Our model accurately localizes both \textbf{scratch} and \textbf{damage} defects.}
\label{fig:real_battery}
\end{figure}

\begin{figure}[h]
\centering
\includegraphics[width=\textwidth]{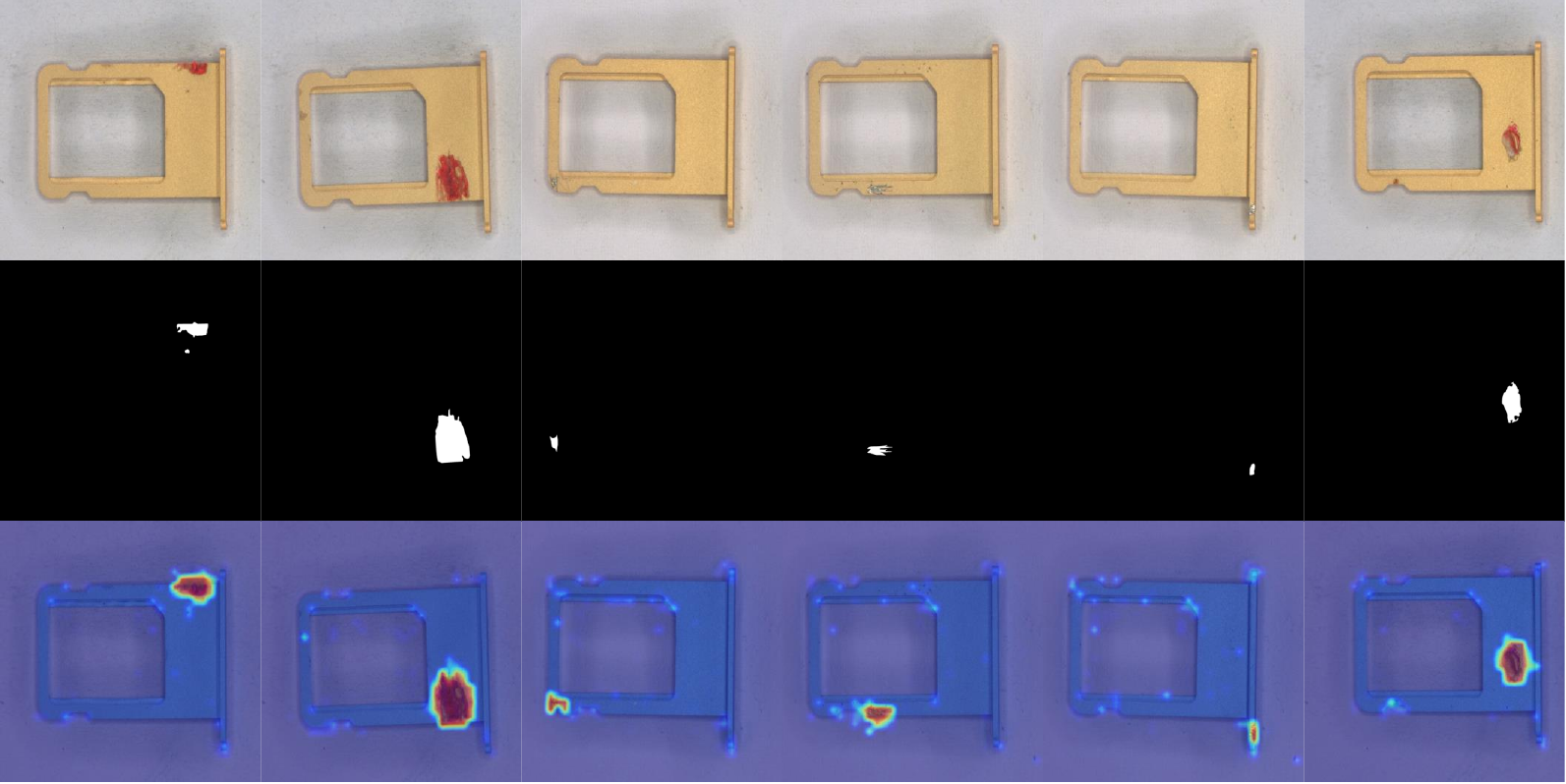}
\caption{The visualization highlights the \textbf{SIM Card Set} object from the Real-IAD dataset. DAPO successfully segments the \textbf{scratch} defect with clear localization.}
\label{fig:real_sim}
\end{figure}

\begin{figure}[h]
\centering
\includegraphics[width=\textwidth]{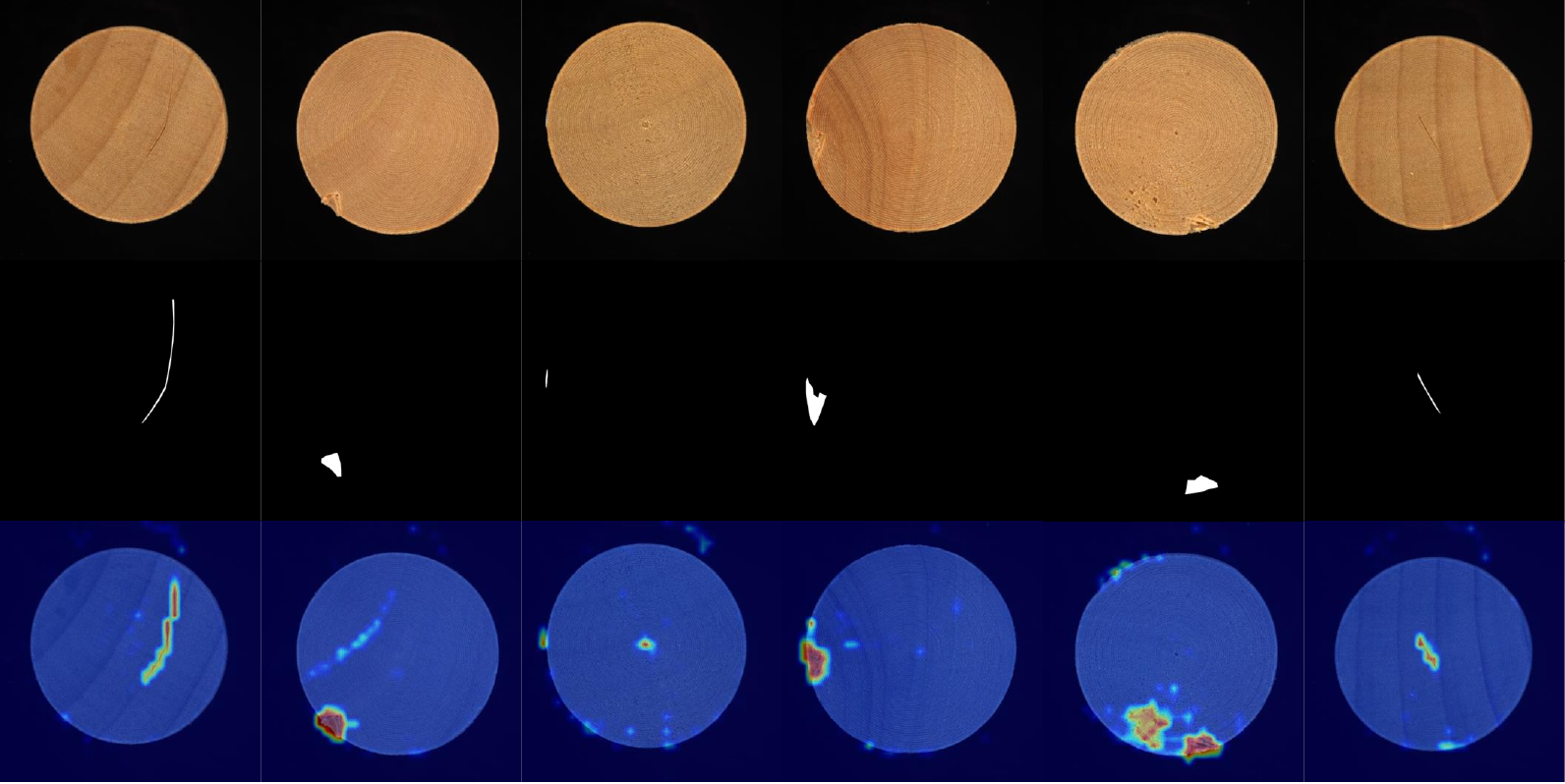}
\caption{The visualization depicts the \textbf{Firewood} object from the Real-IAD dataset. Our model effectively localizes both \textbf{scratch} and \textbf{damage} defects, despite the irregular surface texture.}
\label{fig:real_wood}
\end{figure}

Here, we present the visualization of our anomaly segmentation results. We present six examples of object from VisA, MPDD, MVTec-AD, and Real-IAD datasets. Figures~\ref{fig:visa_capsule}, \ref{fig:visa_fryum}, and~\ref{fig:visa_pcb} shows the capsule, fryum and pcb from VisA. Figures~\ref{fig:mpdd_tubes} and~\ref{fig:mpdd_metalplate} shows the tubes and metal plate from MPDD dataset. Figures~\ref{fig:real_battery}, \ref{fig:real_sim}, and~\ref{fig:real_wood} shows the Phone battery, sim card set, and fire wood from Real-IAD dataset. All visualizations illustrate segmentation results under distribution shift, where the model relies solely on prior knowledge of defect types available in the target set.

\begin{table}[t]
    \centering
    \caption{\textbf{Binary Anomaly Detection and Segmentation:} Fine-grained performance on VisA (top) and MPDD (bottom) datasets.}
    \label{tab:binary_visa_mpdd}
    \small
    \setlength{\tabcolsep}{8pt} 
    \renewcommand{\arraystretch}{1.1}

    \begin{tabular}{lccccc}
        \toprule
        \multicolumn{6}{c}{\textbf{VisA Dataset}} \\
        \textbf{Object} & \textbf{P-AUC} & \textbf{P-PRO} & \textbf{I-AUC} & \textbf{I-AP} & \textbf{I-F1} \\
        \midrule
        Candle & 98.3 & 91.5 & 89.3 & 91.9 & 82.0 \\
        Capsules & 96.6 & 73.1 & 86.3 & 92.5 & 84.4 \\
        Cashew & 88.0 & 79.5 & 85.1 & 94.1 & 85.4 \\
        Chewinggum & 99.5 & 89.5 & 95.9 & 98.4 & 95.3 \\
        Fryum & 92.9 & 84.2 & 92.9 & 96.6 & 91.0 \\
        Macaroni1 & 98.7 & 90.7 & 74.8 & 79.0 & 71.5 \\
        Macaroni2 & 97.0 & 79.6 & 59.5 & 61.1 & 69.1 \\
        PCB1 & 90.3 & 86.2 & 75.1 & 79.4 & 71.2 \\
        PCB2 & 89.2 & 75.0 & 79.4 & 80.5 & 75.1 \\
        PCB3 & 90.4 & 78.3 & 69.4 & 74.6 & 66.9 \\
        PCB4 & 94.8 & 86.9 & 97.0 & 97.1 & 91.9 \\
        Pipe Fryum & 96.4 & 93.7 & 99.8 & 99.9 & 98.5 \\
        \midrule
        \textbf{Mean} & \textbf{94.3} & \textbf{84.0} & \textbf{83.7} & \textbf{87.1} & \textbf{81.9} \\
        \bottomrule
    \end{tabular}

    \vspace{1em} 

    \begin{tabular}{lccccc}
        \toprule
        \multicolumn{6}{c}{\textbf{MPDD Dataset}} \\
        \textbf{Object} & \textbf{P-AUC} & \textbf{P-PRO} & \textbf{I-AUC} & \textbf{I-AP} & \textbf{I-F1} \\
        \midrule
        Bracket Black & 96.5 & 90.3 & 73.6 & 78.9 & 81.5 \\
        Bracket Brown & 91.1 & 74.4 & 59.6 & 77.3 & 80.6 \\
        Bracket White & 97.1 & 87.5 & 85.4 & 83.1 & 81.7 \\
        Connector & 94.7 & 82.9 & 75.2 & 65.0 & 63.4 \\
        Metal Plate & 92.1 & 75.3 & 96.1 & 98.7 & 93.5 \\
        Tubes & 99.0 & 96.1 & 97.0 & 98.8 & 95.5 \\
        \midrule
        \textbf{Mean} & \textbf{95.1} & \textbf{84.4} & \textbf{81.2} & \textbf{83.6} & \textbf{82.7} \\
        \bottomrule
    \end{tabular}
\end{table}

\begin{table}[t!]
    \centering
    \caption{\textbf{Binary Anomaly Detection and Segmentation:} Fine-grained data-subset wise performance on MAD-Sim and MAD-Real datasets.}
    \label{tab:binary_mad_combined}
    \small 
    \setlength{\tabcolsep}{10pt} 
    \renewcommand{\arraystretch}{1.1}
        \begin{tabular}{lccccc}
            \toprule
            \multicolumn{6}{c}{\textbf{MAD-Sim Dataset}} \\
            \textbf{Object} & \textbf{P-AUC} & \textbf{P-PRO} & \textbf{I-AUC} & \textbf{I-AP} & \textbf{I-F1} \\
            \midrule
            Elephant & 67.6 & 57.9 & 61.2 & 91.7 & 94.0 \\
            Owl & 90.0 & 65.6 & 54.9 & 89.4 & 93.0 \\
            Obesobeso & 93.1 & 63.5 & 59.8 & 91.2 & 94.1 \\
            Sabertooth & 88.3 & 56.0 & 53.2 & 87.9 & 93.1 \\
            Zalika & 83.7 & 52.1 & 55.9 & 89.3 & 93.3 \\
            Mallard & 85.2 & 50.0 & 58.1 & 93.7 & 95.6 \\
            Bear & 89.3 & 69.8 & 64.5 & 91.7 & 94.1 \\
            Cat & 92.4 & 56.1 & 48.0 & 89.5 & 94.5 \\
            Pig & 93.3 & 61.6 & 60.0 & 91.7 & 94.0 \\
            Puppy & 85.5 & 60.9 & 60.5 & 90.2 & 92.9 \\
            Parrot & 81.8 & 65.4 & 54.7 & 87.2 & 91.8 \\
            Pheonix & 81.6 & 55.0 & 52.9 & 90.4 & 94.4 \\
            Gorilla & 91.2 & 61.2 & 60.4 & 94.3 & 96.0 \\
            Unicorn & 83.5 & 50.7 & 51.7 & 92.4 & 95.9 \\
            Bird & 89.9 & 47.8 & 55.9 & 92.0 & 94.4 \\
            Swan & 87.2 & 56.4 & 51.9 & 88.5 & 93.3 \\
            Whale & 87.7 & 69.5 & 53.9 & 89.6 & 94.4 \\
            Scorpion & 89.0 & 72.3 & 56.5 & 89.5 & 92.9 \\
            Sheep & 92.8 & 57.9 & 56.9 & 90.6 & 93.7 \\
            Turtle & 89.5 & 61.3 & 76.4 & 97.2 & 95.4 \\
            \midrule
            \textbf{Mean} & \textbf{87.1} & \textbf{59.5} & \textbf{57.4} & \textbf{90.9} & \textbf{94.0} \\
            \bottomrule
        \end{tabular}
        
    \vspace{1em}
    
        \begin{tabular}{lccccc}
            \toprule
            \multicolumn{6}{c}{\textbf{MAD-Real Dataset}} \\
            \textbf{Object} & \textbf{P-AUC} & \textbf{P-PRO} & \textbf{I-AUC} & \textbf{I-AP} & \textbf{I-F1} \\
            \midrule
            Bear & 99.7 & 98.6 & 79.2 & 95.0 & 94.1 \\
            Bird & 78.0 & 40.3 & 54.5 & 85.5 & 89.8 \\
            Elephant & 78.0 & 47.6 & 78.9 & 93.7 & 88.9 \\
            Parrot & 60.7 & 40.1 & 75.7 & 95.1 & 90.2 \\
            Pig & 83.8 & 46.9 & 27.1 & 71.4 & 87.2 \\
            Puppy & 99.6 & 96.4 & 99.0 & 99.8 & 97.6 \\
            Scorpion & 67.8 & 21.5 & 76.5 & 95.0 & 90.2 \\
            Turtle & 99.6 & 97.0 & 100.0 & 100.0 & 100.0 \\
            Unicorn & 83.1 & 52.8 & 68.6 & 91.1 & 89.4 \\
            Whale & 98.8 & 91.4 & 92.5 & 98.8 & 95.2 \\
            \midrule
            \textbf{Mean} & \textbf{84.9} & \textbf{63.3} & \textbf{75.2} & \textbf{92.6} & \textbf{92.3} \\
            \bottomrule
        \end{tabular}
\end{table}

\begin{table}[t!]
    \centering
    \caption{\textbf{Binary Anomaly Detection and Segmentation:} Fine-grained performance on the Real-IAD dataset.}
    \label{tab:realiad}
    \small
    \setlength{\tabcolsep}{12pt} 
    \renewcommand{\arraystretch}{1.05}
    \begin{tabular}{lccccc}
        \toprule
        \textbf{Object} & \textbf{P-AUC} & \textbf{P-PRO} & \textbf{I-AUC} & \textbf{I-AP} & \textbf{I-F1} \\
        \midrule
        Audiojack & 97.3 & 79.5 & 62.2 & 55.1 & 52.5 \\
        VCPill & 98.7 & 89.4 & 90.6 & 88.2 & 78.9 \\
        Switch & 85.6 & 71.1 & 81.3 & 90.2 & 80.7 \\
        USB Adaptor & 94.4 & 57.0 & 76.8 & 73.2 & 69.5 \\
        Eraser & 99.2 & 86.5 & 87.0 & 87.6 & 79.7 \\
        Fire Hood & 99.1 & 93.7 & 95.8 & 90.4 & 85.1 \\
        End Cap & 93.7 & 72.0 & 68.1 & 75.4 & 75.5 \\
        Phone Bat. & 98.4 & 83.4 & 89.1 & 91.9 & 84.9 \\
        Wood Beads & 97.3 & 48.7 & 71.4 & 78.2 & 72.2 \\
        PCB & 95.8 & 78.5 & 79.2 & 85.4 & 81.0 \\
        Mint & 95.3 & 58.2 & 69.9 & 79.0 & 74.8 \\
        Plastic Plug & 98.3 & 85.9 & 88.4 & 88.1 & 79.1 \\
        Transistor & 92.8 & 71.8 & 73.0 & 84.0 & 78.3 \\
        Regulator & 96.2 & 73.6 & 64.9 & 38.8 & 39.2 \\
        Toy & 82.6 & 62.9 & 77.3 & 88.7 & 81.2 \\
        U-Block & 97.9 & 81.7 & 88.2 & 80.5 & 71.7 \\
        Sim Card & 99.5 & 93.7 & 96.0 & 97.3 & 92.7 \\
        Plastic Nut & 95.3 & 67.2 & 84.4 & 64.7 & 56.4 \\
        Button Bat. & 97.0 & 76.3 & 79.4 & 86.4 & 77.7 \\
        Toothbrush & 96.3 & 87.6 & 87.7 & 93.0 & 84.3 \\
        Mounts & 99.7 & 97.6 & 91.6 & 85.5 & 82.7 \\
        Tape & 99.6 & 94.9 & 98.5 & 98.0 & 92.7 \\
        Woodstick & 99.1 & 90.5 & 88.3 & 74.2 & 67.9 \\
        Bottle Cap & 97.9 & 84.8 & 86.4 & 87.2 & 77.8 \\
        Porcelain & 97.4 & 85.1 & 93.1 & 89.7 & 81.2 \\
        Toy Brick & 98.8 & 86.6 & 84.7 & 83.9 & 75.0 \\
        Rolled Strip & 99.2 & 93.8 & 94.3 & 97.4 & 91.3 \\
        Terminal & 97.8 & 84.6 & 94.4 & 96.8 & 91.6 \\
        Zipper & 93.0 & 83.6 & 97.5 & 99.0 & 95.8 \\
        USB & 97.6 & 88.8 & 90.3 & 91.7 & 83.4 \\
        \midrule
        \textbf{Mean} & \textbf{96.4} & \textbf{80.3} & \textbf{84.3} & \textbf{84.0} & \textbf{77.8} \\
        \bottomrule
    \end{tabular}
\end{table}

\begin{table}[t]
    \centering
    \caption{\textbf{Exploratory Image-Level Classification Results (Macro-Average \%).} The high Recall vs. Low Precision trade-off indicates high sensitivity to anomalies but lower specificity for exact defect types.}
    \label{tab:exploratory_visa_mpdd}
    \small
    \setlength{\tabcolsep}{10pt}
    \renewcommand{\arraystretch}{1.1}
        \begin{tabular}{lccc}
            \toprule
            \multicolumn{4}{c}{\textbf{VisA Dataset}} \\
            \textbf{Category} & \textbf{Precision} & \textbf{Recall} & \textbf{F1-Score} \\
            \midrule
            Candle & 8.2 & 99.1 & 15.1 \\
            Capsules & 88.1 & 81.7 & 83.8 \\
            Cashew & 16.6 & 76.1 & 26.6 \\
            Chewinggum & 25.5 & 97.7 & 39.2 \\
            Fryum & 18.4 & 71.1 & 28.6 \\
            Macaroni1 & 11.5 & 99.3 & 20.2 \\
            Macaroni2 & 8.8 & 100.0 & 15.5 \\
            PCB1 & 13.1 & 80.3 & 21.6 \\
            PCB2 & 13.1 & 98.5 & 22.2 \\
            PCB3 & 12.0 & 100.0 & 21.1 \\
            PCB4 & 5.2 & 100.0 & 9.5 \\
            Pipe Fryum & 19.5 & 99.3 & 32.0 \\
            \bottomrule
        \end{tabular}
        
    \vspace{1em}
    
        \begin{tabular}{lccc}
            \toprule
            \multicolumn{4}{c}{\textbf{MPDD Dataset}} \\
            \textbf{Category} & \textbf{Precision} & \textbf{Recall} & \textbf{F1-Score} \\
            \midrule
            Bracket Black & 36.0 & 94.3 & 49.5 \\
            Bracket Brown & 35.3 & 70.6 & 45.5 \\
            Bracket White & 27.4 & 100.0 & 43.0 \\
            Connector & 15.9 & 50.0 & 24.1 \\
            Metal Plate & 24.4 & 100.0 & 38.5 \\
            Tubes & 84.5 & 51.6 & 43.9 \\
            \bottomrule
        \end{tabular}
\end{table}

\begin{table}[t]
    \centering
    \caption{\textbf{Multi-Type Segmentation:} Fine-grained data-subset wise results on MAD dataset.}
    \label{tab:mad_multitype}
    \small
    \setlength{\tabcolsep}{10pt}
    \renewcommand{\arraystretch}{1.1}
        \begin{tabular}{lccc}
            \toprule
            \multicolumn{4}{c}{\textbf{MAD-Sim Dataset}} \\
            \textbf{Object} & \textbf{AUROC} & \textbf{F1} & \textbf{AP} \\
            \midrule
            Bear & 93.7 & 25.0 & 37.6 \\
            Bird & 94.2 & 25.0 & 27.9 \\
            Cat & 94.9 & 25.0 & 27.0 \\
            Elephant & 81.7 & 24.9 & 33.1 \\
            Gorilla & 93.5 & 24.9 & 29.6 \\
            Mallard & 91.4 & 24.9 & 36.8 \\
            Obesobeso & 94.7 & 25.0 & 27.2 \\
            Owl & 93.4 & 24.9 & 34.6 \\
            Parrot & 89.3 & 25.0 & 34.9 \\
            Pheonix & 87.6 & 24.9 & 25.9 \\
            Pig & 95.4 & 24.9 & 32.3 \\
            Puppy & 90.7 & 25.1 & 44.3 \\
            Sabertooth & 93.2 & 24.9 & 30.5 \\
            Scorpion & 92.5 & 25.1 & 32.0 \\
            Sheep & 95.6 & 24.9 & 36.6 \\
            Swan & 92.2 & 24.9 & 27.6 \\
            Turtle & 94.3 & 25.0 & 39.5 \\
            Unicorn & 89.4 & 24.9 & 39.8 \\
            Whale & 92.1 & 25.3 & 38.0 \\
            Zalika & 88.8 & 24.9 & 31.6 \\
            \midrule
            \textbf{Mean} & \textbf{91.9} & \textbf{25.0} & \textbf{33.3} \\
            \bottomrule
        \end{tabular}
        
    \vspace{1em}
        
        \begin{tabular}{lccc}
            \toprule
            \multicolumn{4}{c}{\textbf{MAD-Real Dataset}} \\
            \textbf{Object} & \textbf{AUROC} & \textbf{F1} & \textbf{AP} \\
            \midrule
            Bear & 98.7 & 62.3 & 60.6 \\
            Bird & 80.8 & 49.8 & 51.0 \\
            Elephant & 81.2 & 50.8 & 50.9 \\
            Parrot & 69.4 & 49.8 & 50.4 \\
            Pig & 91.0 & 49.9 & 51.5 \\
            Puppy & 98.8 & 75.1 & 77.0 \\
            Scorpion & 65.0 & 49.6 & 50.6 \\
            Turtle & 99.3 & 74.7 & 72.4 \\
            Unicorn & 78.7 & 49.9 & 50.3 \\
            Whale & 98.0 & 72.4 & 70.8 \\
            \midrule    
            \textbf{Mean} & \textbf{86.1} & \textbf{58.5} & \textbf{58.5} \\
            \bottomrule
        \end{tabular}
\end{table}

\begin{table}[t]
    \centering
    \caption{\textbf{Multi-Type Segmentation on MPDD:} Detailed defect-wise and object-wise performance. (*) indicates defects unseen during training.}
    \label{tab:mpdd_multitype}
    \small
    \setlength{\tabcolsep}{10pt}
    \renewcommand{\arraystretch}{1.1}
        \begin{tabular}{lcc}
            \toprule
            \multicolumn{3}{c}{\textbf{Defect-Wise Results}} \\
            \textbf{Defect Type} & \textbf{AUROC} & \textbf{F1} \\
            \midrule
            Good & 95.66 & 99.86 \\
            Bent & 90.37 & 2.85 \\
            Defective painting* & 93.22 & 0.06 \\
            Flattening* & 98.79 & 66.57 \\
            Hole & 98.62 & 1.14 \\
            Mismatch* & 93.71 & 17.94 \\
            Rust* & 90.12 & 28.37 \\
            Scratch & 97.35 & 26.79 \\
            \bottomrule
        \end{tabular}
        
    \vspace{1em}

        \begin{tabular}{lccc}
            \toprule
            \multicolumn{4}{c}{\textbf{Object-Wise Results}} \\
            \textbf{Object} & \textbf{AUROC} & \textbf{F1} & \textbf{AP} \\
            \midrule
            Bracket black & 97.1 & 37.9 & 35.9 \\
            Bracket brown & 91.1 & 33.4 & 34.6 \\
            Bracket white & 97.0 & 38.5 & 37.0 \\
            Connector & 95.2 & 68.2 & 67.4 \\
            Metal plate & 93.7 & 41.6 & 63.5 \\
            Tubes & 98.8 & 80.2 & 83.3 \\
            \midrule
            \textbf{Mean} & \textbf{95.5} & \textbf{50.0} & \textbf{53.6} \\
            \bottomrule
        \end{tabular}
\end{table}

\begin{table}[t]
    \centering
    \caption{\textbf{Multi-Type Segmentation on VisA:} Detailed defect-wise and object-wise performance. (*) indicates defects unseen during training.}
    \label{tab:visa_multitype}
    
    \small 
    \setlength{\tabcolsep}{10pt} 
    \renewcommand{\arraystretch}{1.2} 
        \begin{tabular}{lcc|lcc}
            \toprule
            \multicolumn{6}{c}{\textbf{Defect-Wise Results}} \\
            \textbf{Defect} & \textbf{AUC} & \textbf{F1} & \textbf{Defect} & \textbf{AUC} & \textbf{F1} \\
            \midrule
            Normal & 86.79 & 99.9 & Melded & 95.97 & 9.51 \\
            Bent & 90.83 & 1.08 & Melt & 93.47 & 7.54 \\
            Breakage & 97.05 & 20.8 & Missing & 87.62 & 13.3 \\
            Bubble & 77.03 & 1.71 & Particle & 99.03 & 3.10 \\
            Burnt & 95.64 & 8.46 & Scratch & 89.73 & 1.89 \\
            Chip & 81.96 & 0.17 & Spot & 87.96 & 1.11 \\
            Crack & 83.75 & 0.14 & Stuck & 86.70 & 9.04 \\
            Damage & 93.24 & 1.63 & Weird wick & 79.04 & 0.05 \\
            Extra & 95.33 & 3.49 & Wrong place & 93.24 & 0.29 \\
            Hole & 95.24 & 1.49 & & & \\
            \bottomrule
        \end{tabular}
        
    \vspace{1em}

        \begin{tabular}{lccc}
            \toprule
            \multicolumn{4}{c}{\textbf{Object-Wise Results}} \\
            \textbf{Object} & \textbf{AUC} & \textbf{F1} & \textbf{AP} \\
            \midrule
            Candle & 90.7 & 13.9 & 16.0 \\
            Capsules & 77.0 & 54.8 & 50.9 \\
            Cashew & 89.6 & 18.4 & 25.0 \\
            Chewinggum & 96.7 & 27.2 & 42.4 \\
            Fryum & 91.4 & 18.6 & 20.9 \\
            Macaroni1 & 88.2 & 20.0 & 20.1 \\
            Macaroni2 & 77.8 & 16.7 & 16.7 \\
            PCB1 & 90.3 & 20.1 & 23.5 \\
            PCB2 & 88.6 & 20.1 & 21.7 \\
            PCB3 & 85.6 & 20.0 & 20.9 \\
            PCB4 & 93.9 & 15.7 & 16.2 \\
            Pipe fryum & 97.0 & 20.9 & 26.3 \\
            \midrule
            \textbf{Mean} & \textbf{88.9} & \textbf{22.2} & \textbf{25.0} \\
            \bottomrule
        \end{tabular}
\end{table}

\begin{figure}[h]
\centering
\includegraphics[width=\textwidth]{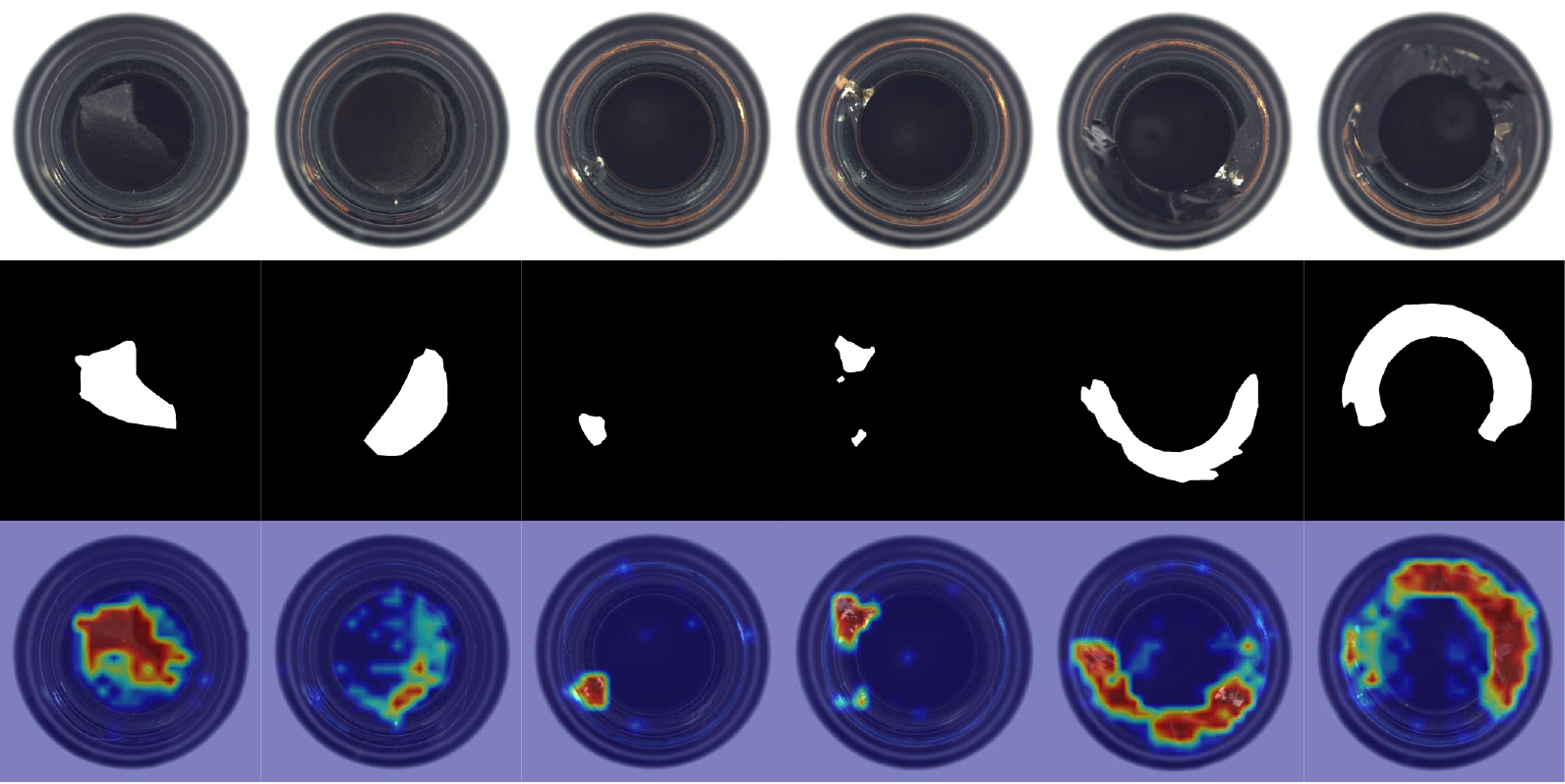}
\caption{The visualization highlights the \textbf{Bottle} object from the MVTec-AD dataset, with the model trained on VisA. Columns 1–2 show localization of the \textbf{contamination} defect, 3–4 capture the \textbf{small break}, while 5–6 successfully segment \textbf{large break}.}
\label{fig:mvtec_bottle}
\end{figure}

